\documentclass[lettersize,journal]{IEEEtran}
\usepackage{amsmath,amsfonts}
\usepackage[english]{babel}
\usepackage{algorithmic}
\usepackage{algorithm}
\usepackage{array}
\usepackage[caption=true,font=small]{subfig}
\usepackage{textcomp}
\usepackage{stfloats}
\usepackage{url}
\usepackage{verbatim}
\usepackage{graphicx}
\usepackage{cite}
\usepackage{balance}
\usepackage{enumitem}
\usepackage{tikz}
\usetikzlibrary {arrows.meta}
\usepackage{amssymb}
\usepackage{multirow}
\usepackage{mathrsfs}
\usepackage{multirow}
\usepackage{hyperref}
\usepackage{xcolor,colortbl}

\usepackage{enumitem}

\graphicspath{{./figures}}

\newcommand{\x}{\mathbf{x}}

\makeatletter
\newcommand{\thickhline}{%
    \noalign {\ifnum 0=`}\fi \hrule height 1pt
    \futurelet \reserved@a \@xhline
}
\newcolumntype{"}{@{\hskip\tabcolsep\vrule width 1pt\hskip\tabcolsep}}
\makeatother

\definecolor{shadow_red}{rgb}{1, 0.9, 0.9}
\definecolor{dark_red}{rgb}{0.8, 0, 0}
\definecolor{mid_red}{rgb}{1, 0.7, 0.7}

\definecolor{shadow_blue}{rgb}{0.9, 0.95, 1}
\definecolor{dark_blue}{rgb}{0, 0.2, 0.8}
\definecolor{mid_blue}{rgb}{0.6, 0.7, 1}

\definecolor{shadow_yellow}{rgb}{1, 0.9, 0.8}
\definecolor{dark_yellow}{rgb}{0.8, 0.5, 0}
\definecolor{mid_yellow}{rgb}{1, 0.8, 0.5}

\definecolor{shadow_purple}{rgb}{0.95, 0.85, 0.9}
\definecolor{dark_purple}{rgb}{0.4, 0.15, 0.25}
\definecolor{mid_purple}{rgb}{0.85, 0.55, 0.7}

\begin{document}

\title{Using Implicit Behavior Cloning and Dynamic Movement Primitive to Facilitate Reinforcement Learning for Robot Motion Planning}

\author{Zengjie Zhang$^{1}$,~\IEEEmembership{Member,~IEEE},
    Jayden Hong$^{2}$,
    Amir M. Soufi Enayati$^{2}$,~\IEEEmembership{Student Member~IEEE},\\
    Homayoun Najjaran$^{2*}$,~\IEEEmembership{Member,~IEEE}
\thanks{This work receives financial support from Kinova\textregistered~Inc. and Natural Sciences and Engineering Research Council (NSERC) Canada under the Grant CRDPJ 543881-19. We also benefit from the hardware resources for the experimental setup provided by Kinova\textregistered~Inc.}
\thanks{$^{1}$Zengjie Zhang is with the Department of Electrical Engineering, Eindhoven University of Technology, Netherlands,  {\tt\small \{z.zhang3\}@tue.nl}.}
\thanks{$^{2}$Jayden Hong, Amir M. Soufi Enayati, and Homayoun Najjaran are with the Faculty of Engineering and Computer Science, University of Victoria, Canada, {\tt\small \{jaydenh, amsoufi, najjaran\}@uvic.ca}.}
\thanks{*Corresponding author.}
}



\maketitle

\begin{abstract}
Reinforcement learning (RL) for motion planning of multi-degree-of-freedom robots still suffers from low efficiency in terms of slow training speed and poor generalizability. In this paper, we propose a novel RL-based robot motion planning framework that uses implicit behavior cloning (IBC) and dynamic movement primitive (DMP) to improve the training speed and generalizability of an off-policy RL agent. IBC utilizes human demonstration data to leverage the training speed of RL, and DMP serves as a heuristic model that transfers motion planning into a simpler planning space. To support this, we also create a human demonstration dataset using a pick-and-place experiment that can be used for similar studies. Comparison studies reveal the advantage of the proposed method over the conventional RL agents with faster training speed and higher scores. A real-robot experiment indicates the applicability of the proposed method to a simple assembly task. Our work provides a novel perspective on using motion primitives and human demonstration to leverage the performance of RL for robot applications.
\end{abstract}


\section{Introduction}\label{sec:intro}

\begin{figure}[htbp]
\centering
\begin{tikzpicture}

\definecolor{shadowgray}{RGB}{105, 105, 105}
\definecolor{darkgray}{RGB}{25, 25, 25}

\definecolor{shadowred}{RGB}{255, 229, 229}
\definecolor{darkred}{RGB}{255, 189, 189}

\node[inner sep=2pt, anchor=north, color=white] (mp) at (-0.1cm, 0cm) {\includegraphics[height=3cm]{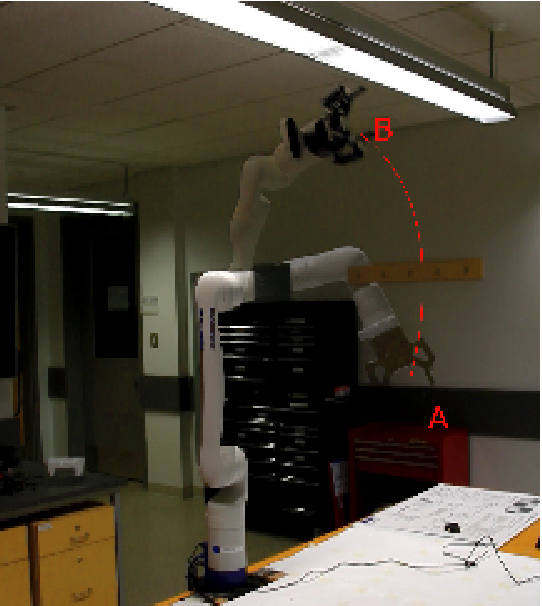}};
\node[anchor=north] () at (mp.south) {\footnotesize{motion planning}};

\node[inner sep=2pt, anchor=north, color=white] (gr) at (2.7cm, 0cm) {\includegraphics[height=3cm]{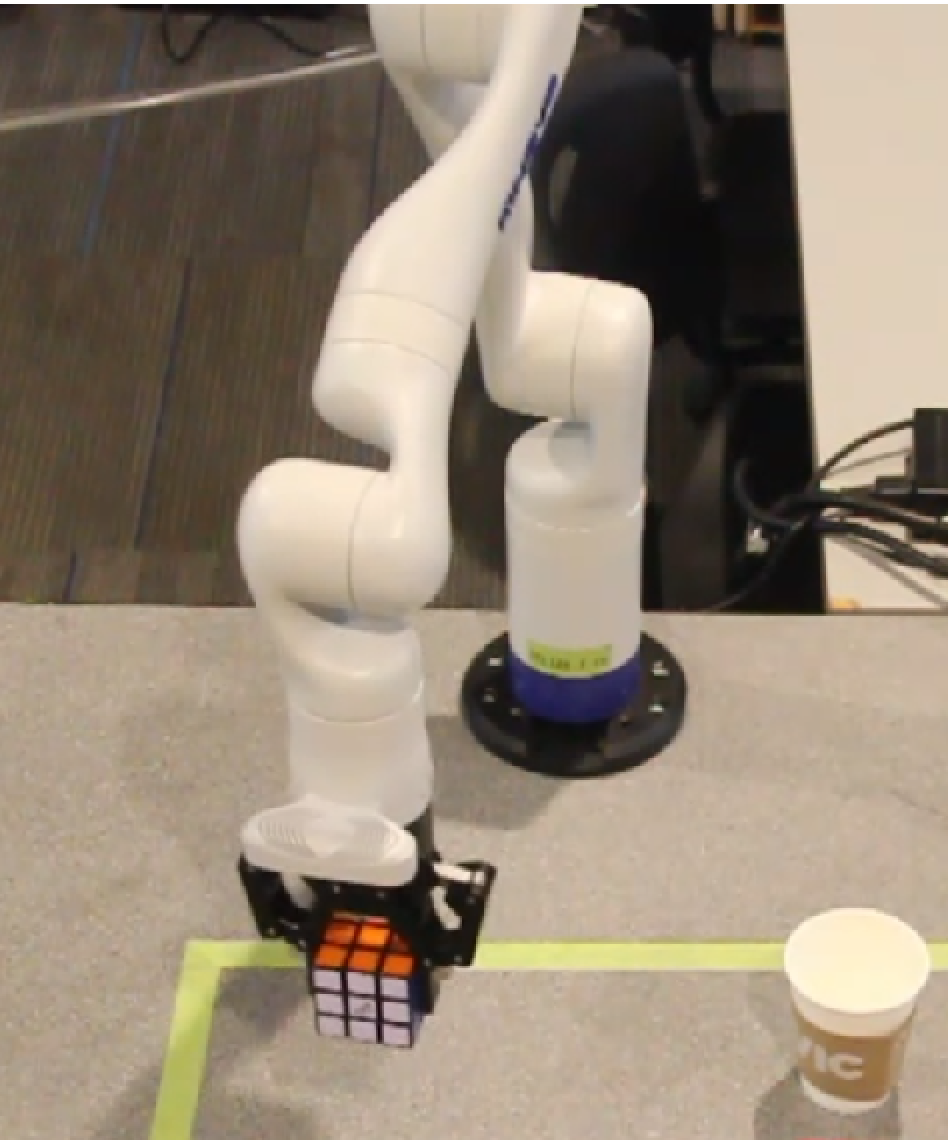}};
\node[anchor=north] () at (gr.south) {\footnotesize{grasping}};

\node[inner sep=2pt, anchor=north, color=white] (am) at (5.4cm, 0cm) {\includegraphics[height=3cm]{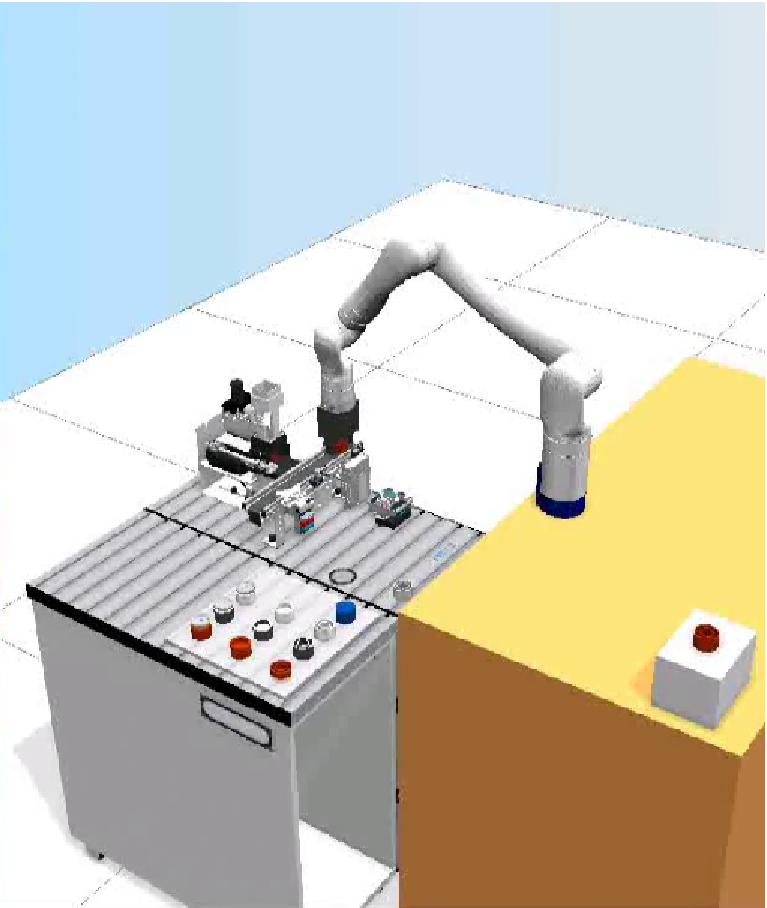}};
\node[anchor=north] () at (am.south) {\footnotesize{manipulation}};

\draw [->, >=Stealth, very thick, black, dotted] plot [smooth, tension=1] coordinates { ([xshift=0.6cm] mp.north) ([xshift=1.3cm,yshift=0.1cm] mp.north) ([xshift=-0.6cm] gr.north)};
\node[minimum height=0.6cm, minimum width=2.6cm, anchor=south] () at ([xshift=1.3cm,yshift=0cm] mp.north) {\footnotesize{\textbf{$+$ gripper motion}}};

\draw[->, >=Stealth, very thick, black, dotted] plot [smooth, tension=1] coordinates { ([xshift=0.6cm] gr.north) ([xshift=1.3cm,yshift=0.1cm] gr.north) ([xshift=-0.6cm] am.north)};
\node[minimum height=0.6cm, minimum width=2.6cm, anchor=south] () at ([xshift=1.3cm,yshift=0cm] gr.north) {\footnotesize{\textbf{$+$ skills}}};

\end{tikzpicture}
     \caption{Motion planning is the essential problem of more complicated robotic tasks, e.g. grasping and manipulation. }
     \label{fig:intro}
\end{figure}

\IEEEPARstart{T}{he} next-generation manufacturing is expected to achieve a higher level of automation with less human power. To this end, intelligent robots are needed to actively learn skills instead of being programmed by experts explicitly~\cite{enayati2022methodical}. Reinforcement learning (RL) is a powerful approach that enables robots to learn an ideal manipulation policy automatically via trial and error. A typical application of RL is robot motion planning which requires the robot to move from an initial position to a goal position without colliding with the obstacles in the environment~\cite{wang2021survey}. As illustrated in Fig.~\ref{fig:intro}, motion planning is an essential problem for more complicated tasks such as grasping, assembly, and manipulation. The conventional approaches used for robot motion planning include optimization-based methods, such as trajectory optimization~\cite{kim2015trajectory} and sequential convex optimization~\cite{schulman2014motion}, and the sampling-based methods, including exploration trees~\cite{perez2020membrane}, probabilistic roadmaps~\cite{ichter2020learned}, and model predictive control~\cite{luis2020online}. These approaches highly depend on precise environmental models. More recent methods, such as stochastic optimization-based planning (STOMP)~\cite{wang2022memory}, attempt to enhance these approaches with data-driven technologies. A review of the conventional motion planning methods can be referred to in~\cite{yang2019survey}. Different from them, RL trains a planner agent automatically via interactions with the environment, instead of directly solving an optimization problem based on a precise environmental model, leading to higher efficiency and flexibility. 

\IEEEpubidadjcol

A typical case of RL is deep reinforcement learning (DRL) which adopts an end-to-end learning scheme using deep neural networks~\cite{arulkumaran2017deep}. It aims at constructing an all-in-one planner for highly-coupled and complicated robot tasks, especially the ones that depend on computer vision, such as grasping~\cite{bao2022learn} and autonomous navigation~\cite{kulhanek2021visual}. Nevertheless, end-to-end learning suffers from the slow convergence rate of the training process and the sensitivity to environmental changes~\cite{enayati2022methodical}. This is due to the large scale of deep neural networks and a high possibility of overfitting. Efforts to resolve these issues include developing high-fidelity simulation platforms~\cite{zhang2022high} or designing sim-to-real schemes to improve its adaptability and robustness~\cite{zhao2020sim}. 
Compared to end-to-end learning, exploiting heuristics can effectively speed up the training of RL agents and avoid overfitting by splitting a big RL problem into several smaller learning problems. Effective heuristic methods for RL include feature extraction layers~\cite{tsurumine2019deep}, modularized neural networks~\cite{cai2021modular}, and hierarchical structures~\cite{yang2021hierarchical}. In~\cite{xiong2020comparison, voigt2020multi}, experimental studies have implied that hierarchical-RL ensures faster convergence and better robustness than end-to-end learning. Besides, transfer learning facilitated by heuristic mappings is used to transfer a simple-task policy to a complicated task without additional training~\cite{da2019survey}. Heuristic models, such as motion primitives and dynamic movement primitives (DMP)~\cite{stulp2011learning} are also widely used to facilitate the design of RL-based robot motion planners by providing a common action space that allows the translation of movements among diverse embodiments.

Behavior cloning (BC) is another technology to promote the training performance of RL agents. The general idea of BC is to force a planning agent to duplicate the human policy encoded in the demonstration data using supervised learning methods~\cite{ly2020learning}. Commonly used by imitation learning (IL) and programming by demonstration (PbD), BC helps robot manipulators to learn human-like policies from demonstrations~\cite{fang2019survey, ravichandar2020recent}. Since human-like policies are widely considered to be decent, it is common to use a BC policy as the initial policy to train an RL agent~\cite{rajeswaran2017learning, tian2021learning, nair2018overcoming}. This method, however, renders complete separation between BC and agent training, in the sense that BC does not help improve the performance of RL. A recent study has proposed a novel method to integrate BC into the training process of an RL agent with a customized training loss, leading to an improvement of the convergence speed~\cite{gupta2021reinforcement}. However, the demonstration used for BC is generated by a PID controller in a simulation environment, instead of real human data. Also, BC is still performed in a \textit{explicit} manner which directly penalizes the deviation between the cloned and the demonstration actions, potentially leading to overfitting to the demonstration. Recent work has attempted to resolve this by proposing an \textit{implicit} BC (IBC) method that penalizes a certain energy function of the cloned policy, such that the cloned policy is less sensitive to the action deviations~\cite{florence2022implicit}. 

We believe that proper usage of both heuristics and human demonstration can leverage the training efficiency and generalizability of RL-based motion planning agents. Although the application of DMP to robot motion planning based on an on-policy RL algorithm has proved to be successful~\cite{stulp2012reinforcement}, whether and how demonstrations can be exploited to promote the training of DMP-based robot planners via an off-policy RL algorithm remains unsolved. Specifically, the following challenging questions need to be answered. 
\begin{itemize}[leftmargin=*]
\item Is it feasible to train DMP using an off-policy algorithm such that the demonstration data can be exploited? 
\item Is BC a decent method to exploit demonstrations for RL?
\item To what extent human demonstrations can promote the performance of an off-policy RL agent? 
\end{itemize}
The answers to these questions may give a novel perspective on designing efficient and generalizable robot motion planners.

To answer these questions, we propose a novel RL method for robot motion planning facilitated by DMP and IBC (IBC-DMP RL), which has not been investigated by existing work, to our best knowledge. The efficacy of the proposed method and its advantages over conventional RL agents are validated using ablation and comparison studies. A hardware experimental study has showcased its applicability to practical robotic tasks. Our detailed contributions are summarized as follows. 
\begin{itemize}[leftmargin=*]
\item We have created a dataset of human motions in a point-to-point reaching (P2PR) task. The dataset is published online and can be openly used for studies on demonstration-facilitated methods for robot motion planning. 
\item We propose a novel IBC-DMP RL framework that improves the learning performance and generalizability of RL-based robot motion planners using demonstrations and DMP.
\item We provide the technical details that are important to ensuring decent training performance of an IBC-DMP RL agent, including \textit{multi-DoF DMP}, \textit{DMP trajectory fitting}, \textit{IBC loss computation}, and \textit{critical loss refinement} which can also inspire the design of other RL-based methods.
\item We conducted comparison studies and hardware experiments to validate the efficacy and advantages of the IBC-DMP RL method and showcase its applicability to practical scenarios.
\end{itemize}
Through these contributions, we give answers to the questions raised above, providing a novel solution for an efficient and generalizable RL-based robot motion planner.

The rest part of the paper is organized as follows. Sec.~\ref{sec:rw} introduces the related work and the preliminary knowledge of the main technologies. Sec.~\ref{sec:frame} interprets the framework of IBC-DMP RL and formulates the problem. Sec.~\ref{sec:human_data} introduces the recording of human trajectories and the generation of demonstrations. Sec.~\ref{sec:training} presents the technical details of IBC-DMP RL. Sec.~\ref{sec:va_sim} and Sec.~\ref{sec:exp} present the comparison and hardware experiments, respectively. Based on a discussion in Sec.~\ref{sec:discussion}, Sec.~\ref{sec:con} concludes the paper.

\textit{Notation}: In this paper, we use $\mathbb{R}$, $\mathbb{R}_{\geq 0}$, and $\mathbb{R}^+$ to represent the sets of real numbers, non-negative real numbers, and positive real numbers, respectively. Also, we use $\mathbb{N}$, $\mathbb{Z}$, and $\mathbb{Z}^+$ to represent the sets of natural numbers, non-negative integers, and positive integers, respectively.

\section{Preliminaries and Related Work}\label{sec:rw}

This section introduces the preliminaries and related work of the three important technologies in this paper, namely DMP, off-policy RL, and BC.

\subsection{Dynamic Movement Primitive (DMP)}\label{sec:dmp}

DMP is a heuristic model commonly used for motion planning of robot manipulators. It has been originally proposed as an abstract model to describe human motions~\cite{schaal2006dynamic}. It is designed as the following virtual second-order linear dynamic model~\cite{stulp2011hierarchical, cohen2021motion},
\begin{subequations}\label{eq:dmp}
\begin{alignat}{2}
\tau \ddot{x}_t = &\,  \alpha \!\left( \beta \!\left( x_{\mathrm{g}} - x_t \right)\! - \dot{x}_t \right)\! + \zeta_t (x_{\mathrm{g}}-x_0 ) f(\zeta_t), \label{eq:dmp_1} \\
\tau \dot{\zeta}_t = &\, - \omega \zeta_t, \label{eq:dmp_2}
\end{alignat}
\end{subequations}
where $x_t, \dot{x}_t, \ddot{x}_t \!\in\! \mathbb{R}$ are the time-dependent position, velocity, and acceleration of the desired trajectory to be generated, respectively, $x_0, x_{\mathrm{g}} \!\in\! \mathbb{R}$ are the initial and the goal positions of the desired trajectory, $\zeta_t \!\in\! \mathbb{R}$ is a canonical dynamic variable with an initial value $\zeta_0\!=\!1$, $\tau \!\in\! \mathbb{R}^+$ is a constant time scalar, $\alpha, \beta, \omega \!\in\! \mathbb{R}^+$ are constant parameters that determine the damping, stiffness, and inertia of DMP, and $f:\mathbb{R}\!\rightarrow\! \mathbb{R}$ is a virtual force function that determines the shape of the generated trajectory. With a proper virtual force function $f$, the DMP model in Eq.~\eqref{eq:dmp} generates a trajectory $x_t$ with smooth position, velocity, and acceleration. Moreover, DMP always ensures a stable trajectory with given initial and goal positions, i.e., the trajectory remains bounded and ultimately converges to the goal position.

DMP can perfectly encode human compliance due to its advantages in smoothness and stability. It also provides a common interface to bridge between human motions and robot trajectories~\cite{kulvicius2011joining}. DMP can also be used to learn an RL policy Without human demonstrations but with a predefined cost function~\cite{hogan2012dynamic}, where the virtual force function is approximated using a radius-basis-function (RBF) neural network and solved using on-policy RL algorithms~\cite{stulp2011hierarchical}. In this way, DMP can easily incorporate collision avoidance constraints into motion planning utilizing a virtual force derived from an artificial potential field (APF)~\cite{li2021reinforcement}.
Thus, DMP-facilitated RL has become a popular method for robot motion planning in recent work~\cite{liang2021dynamic, yuan2022hierarchical}. A survey on the application of DMP to robot manipulation problems is presented in~\cite{saveriano2023dynamic}. 
Nevertheless, DMP is usually used to generate Cartesian-space trajectories, where an inverse-kinematic (IK) algorithm is required~\cite{stulp2012reinforcement}.
Moreover, DMP has been mainly solved using on-policy RL methods, which makes it difficult to promote the current policy using demonstration data. Using off-policy methods to train a DMP model is still an unsolved problem.

\subsection{Off-policy Reinforcement Learning (RL)}\label{sec:mdp}

RL is based on solving a Markov Decision Process (MDP) defined as a tuple $\mathcal{M} \!=\! (\mathcal{S},\mathcal{A}, \mathscr{F}, \mathscr{R}, \gamma)$, where $\mathcal{S}$ is a state space, $\mathcal{A}$ is an action space, $\mathscr{F}: s' \sim p(s'|s, a)$ defines the state transition of the MDP, with $s \in \mathcal{S}$ and $a \in \mathcal{A}$, $\mathscr{R}: \mathcal{S} \times \mathcal{A} \rightarrow \mathbb{R}$ determines the instant reward of the agent, and $0 \!<\!\gamma \!<\!1$ is a discount factor for the reward. RL is dedicated to solving the optimal policy $\pi:a \sim \pi(a|s)$, with $s \in \mathcal{S}$, subject to the following optimization problem,
\begin{equation}\label{eq:optideter}
\textstyle \pi^* = \arg \max_{\pi} \mathbb{E}_{\tau \sim \pi} [R(\tau)],
\end{equation}
where $R(\tau) := \sum_{t=0}^{T} \gamma^{t} \mathscr{R} \!\left(s_t, \pi(s_t) \right)$ is the accumulated reward of a trajectory $\tau := s_0 s_1 \cdots s_T$ with $T \in \mathbb{Z}^+$ and $s_t \in \mathcal{S}$ for all $t \in \{0,1,\cdots,T\}$. The solution $\pi^*$ to Eq.~\eqref{eq:optideter} is referred to as an optimal policy. 

RL methods can be categorized into on-policy and off-policy approaches depending on whether the policy to be promoted is the one that is interacting with the environment. Both types of approaches are widely used for robot motion planning~\cite{wang2021survey, sun2021motion, zhou2022review}. A typical on-policy method is proximal policy optimization (PPO) which updates the policy using the policy gradient calculated from the perturbed system trajectories~\cite{yu2022intelligent}. On-policy methods typically promote policies only after complete episodes. Also, historical policies are no longer used for policy promotion. On the contrary, the policy update of off-policy RL methods, such as deep deterministic policy gradient (DDPG), can be performed at any time. Off-policy methods can also utilize the historical data stored in the experience replay buffer to promote the current policy~\cite{ying2022trajectory}. This means that off-policy methods can learn from multiple policies encoded in the historical data, which is its main advantage over on-policy methods. Nevertheless, the main shortcomings of off-policy methods are long training times and unstable training behaviors due to the bootstrapping effect in the initial stage of the training process. In fact, the training performance of an off-policy agent is highly dependent on the quality of the experience data. In the application of robot motion planning, it is commonly witnessed that on-policy methods outperform off-policy methods when the quality of the experience data is not sufficiently good~\cite{fan2018surreal, naughton2021elastica}. Therefore, many efforts are devoted to improving the performance of off-policy agents by refining the experience data~\cite{hou2017novel, luo2020dynamic}. 

This paper adopts DDPG as the baseline model of the proposed method since it allows for improving the current policy using the experience data. A basic DDPG agent typically consists of two neural networks, namely an actor $\pi_{\theta}(s)$ and a critic $Q_{w}(s,a)$ which are used to approximate the policy and the value function of the agent, respectively, where $\theta$ and $w$ are parameters to be learned. With a trained actor, actions are sampled from a Gaussian distribution $\mathcal{N}(\pi_{\theta}(s), \sigma^2)$ to render a stochastic policy, where $\sigma \in \mathbb{R}^+$ is a hyperparameter to determine the standard deviation of the policy. Details about the definition of the value function can be found in~\cite{hou2017novel}. The main objective of training a DDPG agent is to constantly update the parameters $\theta$ and $w$, such that the approximation $\mathcal{N}(\pi_{\theta}, \sigma^2) \rightarrow \pi^*$ is as close as possible. Another two neural networks $\pi_{\theta'}(s)$ and $Q_{w'}(s,a)$ with parameters $\theta'$ and $w'$, referred to as \textit{target neural networks} are typically used to smooth out the approximation. The parameters $\theta'$ and $w'$ are iteratively updated following the updates of $\theta$ and $w$,
\begin{equation}\label{eq:tac_grad}
\theta' \leftarrow \lambda \theta' + (1- \lambda) \theta,~w' \leftarrow \lambda w' + (1-\lambda) w,
\end{equation}
where $0 \!<\! \lambda \!<\! 1$ is a target factor used to average the updating and stabilize the approximation. 

Training the critic $Q_w$ is a supervised learning process. The samples used to train the network are from an \textit{experience replay buffer} $\mathcal{B}$ sized $n \!\in\! \mathbb{Z}^+$, which is randomly sampled from an experience replay buffer $\mathcal{B}$. Each sample in the buffer $\mathcal{B}$ is organized as the format $\{\,s_j, a_j, s_j', r_j, d_j\}$, where $s_j \in \mathcal{S}$ and $a_j \in \mathcal{A}$ are the state and the action of the agent at a certain history instant $j\in\{1, 2, \cdots\}$, $n$, $s_j' \!\sim\! p(s_j'|s_j, a_j)$ is the \textit{successive state}, $r_j \!=\! \mathscr{R}(s_j, a_j)$ is the corresponding instant reward, and $d_j$ is a boolean termination flag that determines whether an episode terminates or not. The critic loss function of the buffer $\mathcal{B}$ is computed as 
\begin{equation}\label{eq:conv_critic_loss}
\textstyle \mathcal{L}_C(\mathcal{B}) = \frac{1}{n} \sum_{j=1}^n \left(l_j - Q_{w}(s_j, a_j) \right)^2,
\end{equation}
where $l_j = r_j + \gamma (1-d_j) Q_{w'}(s'_j, \pi_{\theta'}(s'_j))$ is the label of sample $(s_j, a_j)$. The actor $\pi_{\theta}$ is also trained using the buffer $\mathcal{B}$ with the following loss function,
\begin{equation}\label{eq:conv_actor_loss}
\textstyle \mathcal{L}_A(\mathcal{B}) =- \frac{1}{n} \sum_{j=1}^n Q_w(s_j, \pi_{\theta}(s_j)).
\end{equation}
Given the computed losses $\mathcal{L}_A$ and $\mathcal{L}_C$, the parameters of the actor and the critic are updated with the following gradient-based law, in an iterative and alternative manner,
\begin{equation}\label{eq:ac_grad}
\Delta \theta =- \alpha \nabla_{\theta} \mathcal{L}_A,~\Delta w =- \alpha \nabla_{w} \mathcal{L}_C,
\end{equation}
where $\Delta \theta$ and $\Delta w$ are the parameter increments at each iteration,  $\alpha \in \mathbb{R}^+$ is the learning rate of the neural networks, and $\nabla_{\theta} \mathcal{L}_A$ and $\nabla_{w} \mathcal{L}_C$ are the gradients of the loss functions to the network parameters. Here, without losing generality, we assume both networks have the same learning rate.

\subsection{Behavior Cloning (BC)}\label{sec:bc}

BC is an important technology of IL to duplicate human policy from demonstrations. It has been widely applied to robot motion planning~\cite{choi2020robotic, tamizi2024end}, and autonomous driving~\cite{ly2020learning}. Here, behavior refers to what \textit{actions} humans tend to take under certain \textit{states} and cloning means learning a new policy to fit human behaviors. BC formulates a supervised learning problem where human actions serve as the ground truth labels of the states. Specifically, given a set of demonstrations that are stored in a buffer $\mathcal{B}$ as defined in Sec.~\ref{sec:mdp}, the source pattern refers to the policy that fits the data $(s_j, a_j)$. Then, the objective of BC is to train a parameterized policy $\pi_{\theta}$, such as a deep neural network (DNN)~\cite{farag2018behavior} or a Gaussian mixture model (GMM)~\cite{chen2017robot}, to clone $\tilde{\pi}$. 
The parameter $\theta$ is updated via a gradient-based law $\Delta \theta = -\alpha_{\theta} \nabla_{\theta} \mathcal{L}_I$,
where $\Delta \theta$ is the parameter increment, $\alpha_{\theta} \!\in\! \mathbb{R}^+$ is the learning rate, and $\nabla_{\theta} \mathcal{L}_I$ is the gradient of an imitation loss function $\mathcal{L}_I$ defined as
\begin{equation}\label{eq:bc_loss_old}
\textstyle \mathcal{L}_I(\mathcal{B}) = \sum_{j=1}^n \mathscr{M}^2\!\left( a_j - \pi_{\theta}(s_j) \right),
\end{equation}
where $\mathscr{M}: \mathcal{A} \!\rightarrow\! \mathbb{R}_{\geq 0}$ is a non-negative function that evaluates the deviation between the two policies, which is commonly the 2-norm of vectors. DMP has also been used to simplify the BC process~\cite{li2023prodmp}. 

BC has proved successful for robots to learn human behaviors~\cite{fang2019survey}, motivating us to combine BC and RL to develop a better robot motion planner, instead of directly using BC to generate human-like trajectories. The conventional BC-facilitated RL methods usually perform a fully separate scheme, i.e., to learn an initial policy with BC before training the agent using nominal RL methods~\cite{rajeswaran2017learning, tian2021learning, nair2018overcoming}. Nevertheless, in such a separate scheme, BC is not fully exploited to promote the training of the RL agent. It also lacks the flexibility of achieving a balance between human likeness and the predefined reward. Recent studies made some attempts to integrate BC into the off-policy RL training using a dual-buffer structure~\cite{gupta2021reinforcement, gupta2022exploiting}, which inspires us to use BC from human demonstration to leverage the training performance of RL. 

However, the conventional BC suffers from an overfitting problem. The success of BC relies on a critical assumption that a unimodal Gaussian policy generates the human demonstration. As a result, it only performs decently when the demonstration encodes unimodal human behaviors, such as one produced by a single human in a certain condition for a certain task. For a demonstration encoding multimodal policies, however, BC may lead to degraded performance. This makes BC difficult to integrate to off-policy RL which combines the demonstration data and the experience replay data into a mixed dataset. Specifically, such a mixed dataset renders a multimodal encoding of both the human behavior and the target policy to be learned. Such a multimodal encoding may lead to unexpected extrapolations among different modals~\cite{florence2022implicit} which can be interpreted as overfitting to multimodal policies. This phenomenon is similar to when an unimodal model is forced to fit a multimodal dataset. It has been addressed that BC performs worse than nominal RL agents in many cases due to overfitting~\cite{kumar2021should}. 

Diffusion policies~\cite{wang2023diffusion} and $k$-means discretized BC~\cite{shafiullah2022behavior} have been proposed to mitigate the overfitting drawback of BC. Nevertheless, these methods require prior knowledge of the distribution model of the multimodal demonstration, which is not always practical for general robot motion planning problems. In this paper, we explore using the \textit{implicit BC (IBC)} technology~\cite{florence2022implicit} to mitigate the overfitting issue for robot motion planning. Different from the conventional BC in Eq.~\eqref{eq:bc_loss_old} which we refer to as \textit{explicit BC (EBC)}, IBC suggests the following loss function~\cite{florence2022implicit},
\begin{equation}\label{eq:ibc_loss_old}
\textstyle \mathcal{L}_I(\mathcal{B}) = \sum_{j=1}^n  E_{\theta}(s_j, a_j), 
\end{equation}
where $E_{\theta}: \mathcal{S} \times \mathcal{A}\!\rightarrow\! \mathbb{R}_{\geq 0}$ is a non-negative energy function of demo data $(s_j, a_j)$ parameterized by $\theta$.

Compared to EBC, IBC generates the target policy through an implicit energy function, instead of explicitly penalizing the deviation between the source and the target policies. We will later show that, by properly designing the energy function that exploits the approximated value function, one can clone a policy without overfitting human demonstrations, leading to a flexible interface to integrate BC into an off-policy RL agent.

\section{Framework and Problem Statement}\label{sec:frame}



The overall framework of IBC-DMP RL based on an adapted \textit{multi-DoF DMP} model is shown in Fig.~\ref{fig:framework}. 
The agent is equipped with a dual-buffer structure that allows learning from both experience and demonstration. 
In the following, we specifically explain the adapted multi-DoF DMP model. Then, we clarify the problem to be solved in this paper.

\begin{figure}[htbp]
\noindent
\hspace*{\fill} 
\begin{tikzpicture}[scale=1,font=\small]

\def\sbheight{0.8cm}
\def\sbwidth{1.6cm}
\def\dev{1.5cm}

\definecolor{closedloop}{RGB}{0, 0, 255}

\definecolor{darkgray}{RGB}{220, 220, 220}
\definecolor{shadowgray}{RGB}{245, 245, 245}
\definecolor{shadowred}{RGB}{255, 229, 229}
\definecolor{shadowyellow}{RGB}{255, 255, 229}
\definecolor{shadowblue}{RGB}{229, 240, 255}
\definecolor{shadowgreen}{RGB}{245, 255, 245}
\definecolor{shadoworange}{RGB}{255, 245, 229}

\node[minimum height=1.4cm, minimum width=7.1cm, draw, thick, dashed, rounded corners=0.2cm] (agent) at (2cm,3.75cm) {};
\node[anchor=north west] (title) at (agent.north west) {~\textbf{Experience Replay Buffer}};

\node[minimum height=0.8cm, minimum width=2.8cm, text width=2.4cm, draw, color=blue, thick, fill=shadowblue, rounded corners=0.2cm, align=center] (dmp1) at (0cm,0.4cm) {Multi-DoF DMP (Sec.~\ref{sec:multi_dmp})};
\node[minimum height=0.8cm, minimum width=2.8cm, text width=2.8cm, draw, color=blue,thick, fill=shadowyellow,  rounded corners=0.2cm, align=center] (human_demo) at (0cm,2cm) {\footnotesize{Human Demonstration} (Sec.~\ref{sec:human_data})};
\node[minimum height=0.8cm, minimum width=2.8cm, text width=2.4cm, draw, color=blue, thick, fill=shadowred, rounded corners=0.2cm, align=center] (demo_data) at (0cm,3.6cm) {Demo Buffer (Sec.~\ref{sec:m_overview})};

\node[minimum height=0.8cm, minimum width=2.8cm, text width=2.4cm, draw, color=red, thick, fill=shadowblue, rounded corners=0.2cm, align=center] (dmp2) at (4cm,0.4cm) {Multi-DoF DMP (Sec.~\ref{sec:multi_dmp})};
\node[minimum height=0.8cm, minimum width=2.8cm, text width=2.8cm, draw, color=red, thick, fill=shadowyellow, rounded corners=0.2cm, align=center] (ibc_dmp) at (4cm,2cm) {IBC-DMP Agent (Sec.~\ref{sec:training})};
\node[minimum height=0.8cm, minimum width=2.8cm, text width=2.4cm, draw, color=red, thick, fill=shadowred, rounded corners=0.2cm, align=center] (interaction) at (4cm,3.6cm) {Replay Buffer (Sec.~\ref{sec:m_overview})};

\node[minimum height=0.2cm, minimum width=0.2cm] (plus) at (2cm,3.6cm) {$\mathbf{+}$};

\draw[->,>=stealth,thick, color=blue,] (dmp1.east) -- ([xshift=0.3cm] dmp1.east) -- node[pos=0.5, anchor=east]{$\mathcal{X}_t\!$} ([xshift=0.3cm] human_demo.east) -- (human_demo.east);
\draw[<-,>=stealth,thick, color=blue,] (dmp1.west) -- ([xshift=-0.3cm] dmp1.west) -- node[pos=0.5, anchor=west]{$\tilde{\mathbf{f}}(\mathcal{X}_t)$} ([xshift=-0.3cm] human_demo.west) -- (human_demo.west);

\draw[->,>=stealth,thick, color=blue, dotted] ([xshift=0.3cm] human_demo.east)  -- node[pos=0.5, anchor=east]{$s_t$} ([xshift=0.3cm] demo_data.east) -- (demo_data.east);
\draw[->,>=stealth,thick, color=blue, dotted] ([xshift=-0.3cm] human_demo.west) -- node[pos=0.5, anchor=west]{$a_t$} ([xshift=-0.3cm] demo_data.west) -- (demo_data.west);

\draw[->,>=stealth,thick,color=red] (dmp2.east) -- ([xshift=0.3cm] dmp2.east) -- node[pos=0.5, anchor=east]{$\mathcal{X}_t\!$} ([xshift=0.3cm] ibc_dmp.east) -- (ibc_dmp.east);
\draw[<-,>=stealth,color=red, thick] (dmp2.west) -- ([xshift=-0.3cm] dmp2.west) -- node[pos=0.5, anchor=west]{$\mathbf{f}(\mathcal{X}_t)$} ([xshift=-0.3cm] ibc_dmp.west) -- (ibc_dmp.west);

\draw[->,>=stealth,color=red, thick, dotted] ([xshift=0.3cm] ibc_dmp.east)  -- node[pos=0.5, anchor=east]{$s_t$} ([xshift=0.3cm] interaction.east) -- (interaction.east);
\draw[->,>=stealth,color=red, thick, dotted] ([xshift=-0.3cm] ibc_dmp.west) -- node[pos=0.5, anchor=west]{$a_t$} ([xshift=-0.3cm] interaction.west) -- (interaction.west);

\draw[->,>=stealth,thick, dashed] (agent.south) -- ([yshift=-0.2cm] agent.south) -- ([xshift=0.3cm, yshift=-0.2cm] ibc_dmp.south);

\node[color=blue, anchor=north] () at (0cm,0cm) {Once before training};
\node[color=red, anchor=north] () at (4cm,0cm) {Repeated in training};

\end{tikzpicture}
\hspace{\fill} 
\caption{The IBC-DMP RL framework, where $\mathcal{X}_t$ is the comprehensive state of the multi-DoF DMP model, $\tilde{\mathbf{f}}(\mathcal{X}_t)$ and $\mathbf{f}(\mathcal{X}_t)$ are the virtual force functions provided by the human demonstration and the IBC-DMP agent, respectively, and $s_t$ and $a_t$ are the state and action data stored in the buffers. The solid arrows denote the interactions, the dotted arrows indicate data storing, and the dashed arrow represents agent training. Besides, the left column (in blue) is prepared only once before the training process, while the right column (in red) is repeatedly updated during training.}
\label{fig:framework}
\end{figure}

\subsection{Motion Planning Using Multi-DoF DMP}\label{sec:multi_dmp}

As introduced in Sec.~\ref{sec:dmp}, the advantage of DMP is generating sufficiently smooth and inherently stable trajectories that can be reproduced by general robot platforms with similar position-based control interfaces~\cite{schaal2006dynamic}. Therefore, we use a DMP model as the environment in the robot motion planning problem in this paper.
The conventional DMP in Sec.~\ref{sec:dmp} is defined for a one-dimensional point and has been used for robot motion planning in a decoupled manner, i.e., one DMP is used to generate the trajectory for each DoF of the robot~\cite{stulp2012reinforcement}. This leads to the lack of coupling among different DoFs, which restricts the flexibility of solving the optimal planning policy. To resolve this issue, in this paper, we propose the following multi-DoF DMP model for the motion planning of a robot end-effector in the Cartesian space,
\begin{equation}\label{eq:dmp_multi_dof}
\tau\ddot{\x}_t \!=\! K_{\alpha} \!\left( K_{\beta} \!\left( \x_{\mathrm{g}} \!-\! \x_t \right)\! -\! \dot{\x}_t \right)\! + \zeta_t \|\x_{\mathrm{g}}\!-\!\x_{\mathrm{i}} \| \mathbf{f}(\mathcal{X}_t),
\end{equation}
where $\x_t$, $\dot{\x}_t$, $\ddot{\x}_t \!\in\! \mathbb{R}^3$ are the position, linear velocity, and acceleration of the end-effector in the Cartesian space, $\x_{\mathrm{i}}$, $\x_{\mathrm{g}} \!\in\! \mathbb{R}^3$ are the initial and goal positions of the end-effector, $K_{\alpha}$, $K_{\beta} \!\in\! \mathbb{R}^{3 \times 3}$ are parametric matrices, $\tau \!\in\! \mathbb{R}$ and $\zeta_t \in \mathbb{R}$ are the temporal scalar and the canonical variable, the same as \eqref{eq:dmp}, $\mathcal{X}_t$ is a comprehensive state vector to be determined, and $\mathbf{f}:\mathbb{R}^{10} \!\rightarrow\! \mathbb{R}^3$ is a multi-dimension virtual force function. The initial state of model \eqref{eq:dmp_multi_dof} is set as $\x_0 \!=\! \x_{\mathrm{i}}$ and $\dot{\x}_0 \!=\! 0$.

For brevity, we only consider a \textit{single static obstacle}, a vertically positioned \textit{cylinder}. The position of the obstacle $\x_{\mathrm{b}} \!\in\! \mathbb{R}^3$ is a constant vector assigned as the Cartesian coordinate of the center of its top surface.
Thus, we define the comprehensive state $\mathcal{X}_t$ as a ten-dimensional vector $\mathcal{X}_t\!=\![\,\x_t^{\top}, \dot{\x}_t^{\top}, \x_t^{\top}-\x_{\mathrm{b}}^{\top}, \zeta_t]^{\!\top\!} \!\in\!\mathbb{R}^{10}$.
Here, we encode a time-variant vector $\x_t - \x_{\mathrm{b}}$ into $\mathcal{X}_t$ instead of the constant vector $\x_{\mathrm{b}}$ since the former provides larger diversity to the data. The incorporation of more complicated environments with multiple dynamic obstacles is beyond the scope of this paper and will be considered in future work.
The multi-DoF DMP is different from the conventional DMP~\eqref{eq:dmp} in three aspects. 
\begin{itemize}[leftmargin=*]
\item The virtual force function $\mathbf{f}$ depends not only on the canonical variable $\zeta_t$ but also on the internal states of the DMP, namely $\x_t$ and $\dot{\x}_t$, and the obstacle position $\x_{\mathrm{b}}$. Different from the conventional DMP models, this may change the poles or the closed-loop dynamics of the multi-DoF DMP, which can be both an advantage and a drawback. On the one hand, the state-dependent virtual force function $\mathbf{f}$ can be learned to automatically adjust the poles of the DMP model, such that its dynamic properties are not completely determined by the hyperparameters $K_{\alpha}$ and $K_{\beta}$. This improves the flexibility of agent training. On the other hand, the DMP model may lose its inherent stability. In this paper, we set action limits $\underline{f} \leq \mathbf{f}(\mathcal{X}_t) \leq \overline{f}$ to avoid this.
\item The gain of the virtual force function is the absolute distance between the initial position $\x_{\mathrm{i}}$ and the goal position $\x_{\mathrm{g}}$, $\|\x_{\mathrm{g}}-\x_{\mathrm{i}}\|$, instead of the element-wise distance used by the conventional DMP~\eqref{eq:dmp}. Such a scheme can improve the flexibility of a DMP model by fully incorporating the coupling of its different dimensions.
\item The obstacle position $\x_{\mathrm{b}}$ is included in the state $\mathcal{X}_t$ of the DMP, instead of being incorporated as an additional virtual force term as the conventional DMP model. This provides the possibility of incorporating more complicated obstacle information into motion planning.
\end{itemize}

We discretize the multi-DoF DMP model in Eq. \eqref{eq:dmp_multi_dof} using a sampling time $\Delta t \!\in\! \mathbb{R}^+$ as follows, 
\begin{equation}\label{eq:dmp_multi_dof_dis}
\begin{split}
&\x_{t+1} \!=\! \x_t + \Delta t \dot{\x}_t ,~\dot{\x}_{t+1} \!=\! \dot{\x}_{t} + \Delta t\ddot{\x}_{t}, \\
&\zeta_{t+1} \!=\! \zeta_t - \frac{\Delta t}{\tau} \omega \zeta_t,
\end{split}
\end{equation}
where $\ddot{\x}_t$ is given by \eqref{eq:dmp_multi_dof}.
Then, the generated trajectory is a set of waypoints with a time series $\{0, 1, \cdots, T\}$, with $T \!\in\! \mathbb{Z}^+$ being the length of the trajectory. 

\subsection{Motion Planning Problem with Shaped Reward}\label{sec:statement}

This paper solves a general robot motion planning problem, i.e., given an initial position $\x_{\mathrm{i}} \!\in\! \mathbb{R}^3$ and a goal position $\x_{\mathrm{g}} \!\in\! \mathbb{R}^3$ in the Cartesian space, generate a smooth trajectory $\x_t \!\in\! \mathbb{R}^3$ for $t \in \{0, 1, \cdots, T\}$ using the multi-DoF DMP model in Eq.~\eqref{eq:dmp_multi_dof}, such that $\x_T$ is sufficiently close to $\x_{\mathrm{g}}$. 
The critical point is to solve an optimal virtual force function $\mathbf{f}$ for Eq.~\eqref{eq:dmp_multi_dof} such that a certain reward is maximized. 
This problem renders decision-making over an MDP $\mathcal{M} \!=\! (\mathcal{S},\mathcal{A}, \mathscr{F}, \mathscr{R}, \gamma)$ defined in Sec.~\ref{sec:mdp}, where $\mathcal{S} \!=\!\{\mathcal{X}_t|\x_{\mathrm{i}} \!\in\! \mathbb{R}^3, \, 0\!\leq\! t \!<\! T\}$ is the state space, $\mathcal{A} \!=\!\left\{\mathbf{f}(s) \left| \forall\, s \in \mathcal{S} \right. \right\}$ is the action space, $\mathscr{F}$ is the state transition characterized by \eqref{eq:dmp_multi_dof_dis}, and $\mathscr{R}$ assigns certain rewards to be determined later. In this sense, the policy $\pi$ can be represented as a parameterized virtual force function $\mathbf{f}_{\theta}$, modeled as a neural network and learned using an off-policy RL method introduced in Sec.~\ref{sec:mdp}. Besides, BC introduced in Sec.~\ref{sec:bc} can be used to promote the training of the RL agent with the demonstration policy $\tilde{\mathbf{f}}$ (Fig.~\ref{fig:framework}). Different from the conventional IL problems, this paper does not aim at using $\mathbf{f}_{\theta}$ to approximate the demonstration policy $\tilde{\mathbf{f}}$. Instead, $\tilde{\mathbf{f}}$ is only used to improve the training of $\mathbf{f}_{\theta}$.

The reward $\mathscr{R}$ should be shaped in a way that any generated trajectory ensures sufficient smoothness (minimal acceleration) and safety (a minimal distance from the obstacle). These requirements are encoded in the following cost function, 
\begin{equation}\label{eq:cost}
\mathcal{J}_t = \left\{ \begin{array}{ll}
\sum_{i=1}^4 \alpha_i \mathcal{J}_t^{(i)},  & t < T, \\
\alpha_5 \mathcal{J}_t^{(5)}, & t=T,
\end{array} \right.
\end{equation}
where $\alpha_i \!\in\! \mathbb{R}^+$ are constant parameters, $i=1,2,\cdots,5$, and $\mathcal{J}_t^{(i)} \!\in\! \mathbb{R}$ are instant costs at time $t$, defined as
\begin{equation*}
\begin{split}
&\mathcal{J}_t^{(1)} \!=\! \!\left\|\ddot{\x}_t \right\|^2,~\mathcal{J}_t^{(2)} \!=\! \left\|\x_t \!-\!\x_{\mathrm{g}} \right\|^2, \\
&\mathcal{J}_t^{(3)} \!=\! \left\{ \begin{array}{ll} \!\!\!\eta ( \|\x_t^{(1,2)} \!-\! \x_{\mathrm{b}}^{(1,2)} \| \!-\! r_{\mathrm{b}},\varepsilon_0^{\mathrm{b}},\varepsilon_1^{\mathrm{b}} ) &\mathrm{if}~0\!<\!\x_t^{(3)}\!<\!\x_{\mathrm{b}}^{(3)} \\
\!\!\!0 & \mathrm{otherwise}, \\
\end{array} \right. \\
&\mathcal{J}_t^{(4)} = \eta(\x_t^{(3)},\varepsilon_0^{\mathrm{d}},\varepsilon_1^{\mathrm{d}}),~ \mathcal{J}_t^{(5)} = \xi(\|\x_t \!-\! \x_{\mathrm{g}}\|, \varepsilon_T),
\end{split}
\end{equation*}
where $\x_t^{(1,2)} \!\in\!\mathbb{R}^2$, $\x_t^{(3)} \!\in\!\mathbb{R}$, $\x_{\mathrm{b}}^{(1,2)} \!\in\!\mathbb{R}^2$, $\x_{\mathrm{b}}^{(3)} \!\in\!\mathbb{R}$ are the slides of vectors $\x_t$ and $\x_{\mathrm{b}}$, $r_{\mathrm{b}} \in \mathbb{R}^+$ is the radius of the cylinder, $\eta:\mathbb{R}\!\rightarrow\!\mathbb{R}_{\geq 0}$ with constant parameters $0 \!<\! \varepsilon_0 \!<\! \varepsilon_1$ is an APF function defined as
\begin{equation}
\eta(x, \varepsilon_0, \varepsilon_1) \!=\! \left\{ \begin{array}{ll}
\!\!\!(x\!-\!\varepsilon_0)^{-2} \!-\! (\varepsilon_1 \!-\! \varepsilon_0)^{-2}, &\!\! \varepsilon_0 \!<\! x \!<\! \varepsilon_1, \\
\!\!\!0, &\!\! x \!\geq\! \varepsilon_1,
\end{array} \right.
\end{equation}
$0 \!<\! \varepsilon_0^{\mathrm{b}} \!<\! \varepsilon_1^{\mathrm{b}}$ and $0 \!<\! \varepsilon_0^{\mathrm{d}} \!<\! \varepsilon_1^{\mathrm{d}}$ are APF parameters for the obstacle and the ground, and $\xi:\mathbb{R}\!\rightarrow\!\mathbb{R}_{\geq 0}$ is a squared dead-zone function defined as
\begin{equation}
\xi(x, \varepsilon) = \left\{ \begin{array}{ll}
0, & 0 \leq x \leq \varepsilon, \\
x - \varepsilon, & x > \varepsilon,
\end{array} \right.
\end{equation}
where $\varepsilon \in \mathbb{R}^+$ is the dead-zone scalar. Here, $\eta(x)$ serves as an APF function that penalizes $x$ if it gets close to $\varepsilon_0$ from the positive direction. Another parameter $\varepsilon_1$ limits the domain of impact of $\eta(x)$. To avoid undefinition, we assign $\eta(x)$ with a large number when $x \!\leq\! \varepsilon_0$.

In this sense, the instant reward is defined as $\mathscr{R}_t: - \mathcal{J}_t$. In the comprehensive cost $\mathcal{J}_t$ as Eq.~\eqref{eq:cost}, term $\mathcal{J}_t^{(1)}$ penalizes the value of the acceleration to ensure sufficient smoothness. Term $\mathcal{J}_t^{(2)}$ penalizes the distance between the current position $\x_t$ and the goal position $\x_{\mathrm{g}}$, aiming at fast and straightforward goal reaching. Terms $\mathcal{J}_t^{(3)}$ and $\mathcal{J}_t^{(4)}$ attempt to keep the position $\x_t$ away from the ground ($z=0$ plane) and the cylinder obstacle, respectively. The usage of APF functions ensures a very large cost when $\x_t$ runs into the cylinder obstacle or the ground. The last term $\mathcal{J}_T^{(5)}$ penalizes the distance between the ultimate position $\x_T$ and the goal position $\x_{\mathrm{g}}$. A dead-zone scalar $\varepsilon_T$ is involved to prescribe the tolerable level of the ultimate error. Specifically, the ultimate cost is exerted only when the ultimate error exceeds the threshold $\varepsilon_T$.
By minimizing the comprehensive cost~\eqref{eq:cost} for the multi-DoF DMP model~\eqref{eq:dmp_multi_dof}, one can get a smooth trajectory $\x_t$ that ultimately reaches the goal position $\x_{\mathrm{g}}$ at $t=T$ while avoiding collision with the ground and the obstacle.

The two main objectives of this paper are as follows. (1). Translate human demonstration into data that are compatible with the demo buffer, corresponding to the left column of Fig.~\ref{fig:framework}; (2) Train the optimal policy $\mathbf{f}$ using the demo buffer, as the right column of Fig.~\ref{fig:framework}. The solutions to these two problems will be presented in Sec.~\ref{sec:human_data} and Sec.~\ref{sec:training}, respectively.

\section{Human Demonstration Collection}\label{sec:human_data}

This section interprets the experiments to collect human hand trajectories and translate them into demonstration data. The translated data fit a multi-DoF DMP model which allows them to be reproduced by general robot platforms. The dataset is published at https://doi.org/10.5281/zenodo.11237258. The recorded human motion data are only for public educational purposes. No private information about the subject is disclosed. Also, the configuration of the data recording experiment does not cause any risk of harm to the subject. Therefore, this research is exempt from human-related ethical issues.

\subsection{Human Data Acquisition}\label{sec:hdr}

We use a data-recording experiment to collect human-hand motion in a P2PR task since it is a typical scenario of robot motion planning. The experiment is designed to capture how the human hand attempts to reach a goal position while avoiding collisions with a mid-way obstacle. An HTC VIVE tracking system, consisting of several motion trackers and a pair of virtual reality (VR) base stations is used to record the position of any object attached to the trackers. The system can track the Cartesian pose of the rigid body in a stable sampling frequency $100\,$Hz. The tracking precision can be as high as $1\,$mm.

\begin{figure}[htbp]
     \centering
     
\begin{tikzpicture}

\definecolor{shadowgray}{RGB}{105, 105, 105}
\definecolor{darkgray}{RGB}{25, 25, 25}

\usetikzlibrary{shapes.misc, positioning}

    \node[draw,thick,minimum width=4cm,minimum height=2cm, thick] (table) at (1.5cm, 2.25cm) {};
    \node[anchor=north west] () at(table.north west) {\small desk};

    \node[draw,color=shadowgray,minimum width=1.8cm,minimum height=1.5cm, line width=5pt] (rect) at (2cm, 2.1cm) {};
    \node[circle,inner sep=0pt,minimum size=7mm,fill=darkgray] (human) at (0.7cm, 0.7cm) {};
    \node[rounded rectangle, minimum width=1.6cm, minimum height=0.4cm, fill=darkgray] (torso) at (human.center) {};
    \node[rounded rectangle, rotate=20, minimum width=1.4cm, minimum height=0.3cm, fill=darkgray] (torso) at ([xshift=0.8cm, yshift=0.1cm] human.center) {};
    \node[rounded rectangle, rotate=160, minimum width=1.4cm, minimum height=0.3cm, fill=darkgray] (torso) at ([xshift=0.8cm, yshift=0.45cm] human.center) {};
    \node[anchor=west] () at([xshift=0.8cm, yshift=-0.2cm] human.center) {\small human subject};

    \node[rounded rectangle, rotate=150, minimum width=1.4cm, minimum height=0.3cm, fill=darkgray] (torso) at ([xshift=-0.9cm, yshift=0.2cm] human.center) {};
    \node[rounded rectangle, rotate=30, minimum width=1.4cm, minimum height=0.3cm, fill=darkgray] (torso) at ([xshift=-0.9cm, yshift=0.65cm] human.center) {};
    
    \node[circle,inner sep=0pt,minimum size=1mm,draw, fill=black] (origin) at (-2, 1.5) {};
    \draw[->, >=Stealth,color=red, very thick] (origin.center) --node[pos=1, anchor=north]{$x$} ([xshift=1cm] origin.center);
    \draw[->, >=Stealth,color=blue, very thick] (origin.center) -- node[pos=1, anchor=east]{$y$} ([yshift=1cm] origin.center);

    \node[circle,inner sep=0pt,minimum size=1mm, draw, fill=black] (origin1) at ([xshift=0.1cm, yshift=0.1cm] rect.south west) {};
    \node[anchor=south east] () at([xshift=0cm, yshift=0cm] origin1.center) {\scriptsize $\mathcal{I}$};

    \node[circle,inner sep=0pt,minimum size=3mm, draw, fill=black] (obst) at (1.8cm, 2cm) {};
    \node[anchor=south east] () at(obst.center) {\scriptsize $\mathcal{O}$};

    \node[circle,inner sep=0pt,minimum size=3mm, draw] (target) at (2.4cm, 2.4) {};
    \node[anchor=south east] () at(target.center) {\scriptsize $\mathcal{G}$};

    \draw [->, >=Stealth, very thick, black, dotted] plot [smooth, tension=0.8] coordinates { (origin1.center) (1.8cm,1.5cm) (target.center)};

    \node[inner sep=2pt, anchor=south west, color=white, rotate=-25] (camera1) at (-1.7cm, 3.0cm)
    {\includegraphics[width=0.8cm]{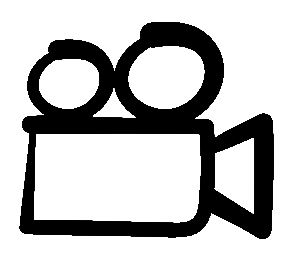}};

    \node[inner sep=2pt, anchor=north west, color=white, rotate=-25, xscale=-1] (camera2) at (4.8cm, 1.2cm])
    {\includegraphics[width=0.8cm]{camera.eps}};

    \node[align=center, anchor=north, text width=1.5cm] () at (camera1.south) {\small base station 1};

    \node[align=center, anchor=south, text width=1.5cm] () at ([xshift=-0.2cm] camera2.north) {\small base station 2};

    \node[anchor=south] () at([xshift=0cm, yshift=-0.14cm] rect.north) {\small motion region};

\end{tikzpicture}
     \caption{The top view of the data recording experiment, where $\mathcal{I}$, $\mathcal{O}$, and $\mathcal{G}$ are the initial, obstacle, and goal positions, respectively.}
     \label{fig:human_exp}
\end{figure}

The top view of the experiment configuration is shown in Fig.~\ref{fig:human_exp}. A human subject sits in front of a plat desk. The desk is $0.35\,$m vertically higher than the seat, aligned with the subject's stomach level such that the subject is seated comfortably. A rectangular planar motion region is marked on the desk with height $0.5\,$mm and width $0.6\,$m which are slightly shorter than the length of an arm of an adult, shown as a thick-line gray box in Fig.~\ref{fig:human_exp}. The motion region is aligned with the desk edge towards the subject. The subject's seat is placed such that the hand can be comfortably positioned in the left-bottom corner of the motion region and can naturally reach any point within the motion region without standing up. Two VR base stations are placed on the diagonal corners of the workspace for superior tracking precision, as shown as camera symbols in Fig.~\ref{fig:human_exp}. The coordinate of the tracking system is calibrated such that its origin coincides with the left-bottom corner of the workspace and its $x$-, $y$-, and $z$-axis point to the right, front, and top of the subject, respectively. We set the initial position $\mathcal{I}$ such that the hand is comfortably placed at the origin. For a given goal $\mathcal{G}$ and an obstacle $\mathcal{O}$, the subject is required to move the hand from the initial position $\mathcal{I}$ to the goal $\mathcal{G}$ naturally while avoiding the obstacle $\mathcal{O}$. A possible hand trajectory is illustrated as a dashed arrow in Fig.~\ref{fig:human_exp}.

\begin{figure}[htbp]
    \centering{
    \subfloat[]{\includegraphics[width=0.155\textwidth]{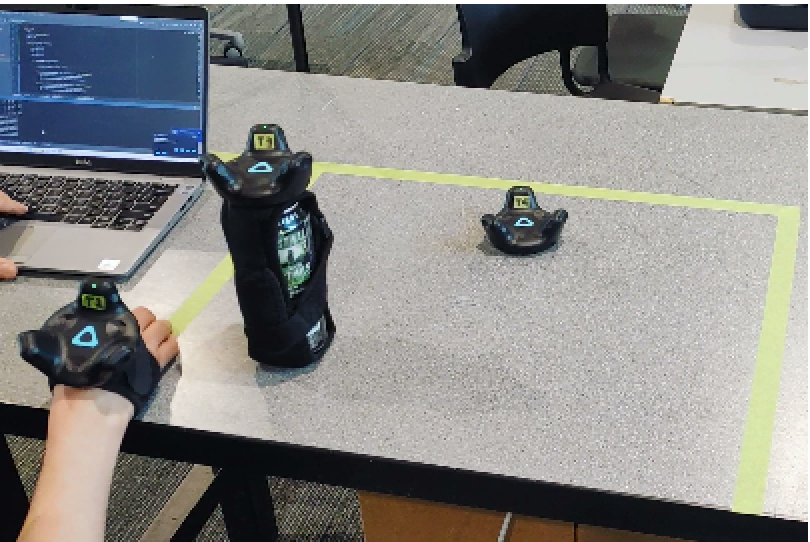}
        \label{exp_obs_1}}
    \subfloat[]{\includegraphics[width=0.155\textwidth]{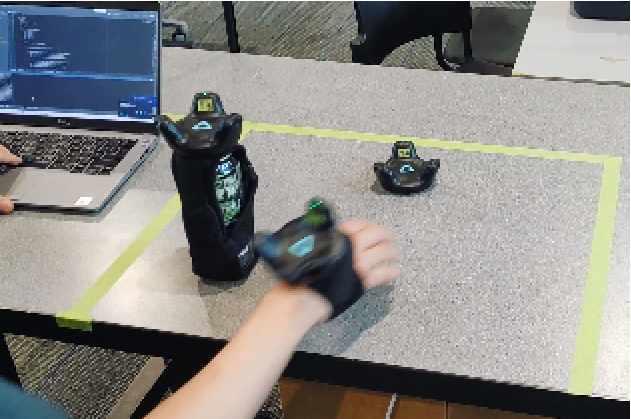}
        \label{exp_obs_3}}
    \subfloat[]{\includegraphics[width=0.155\textwidth]{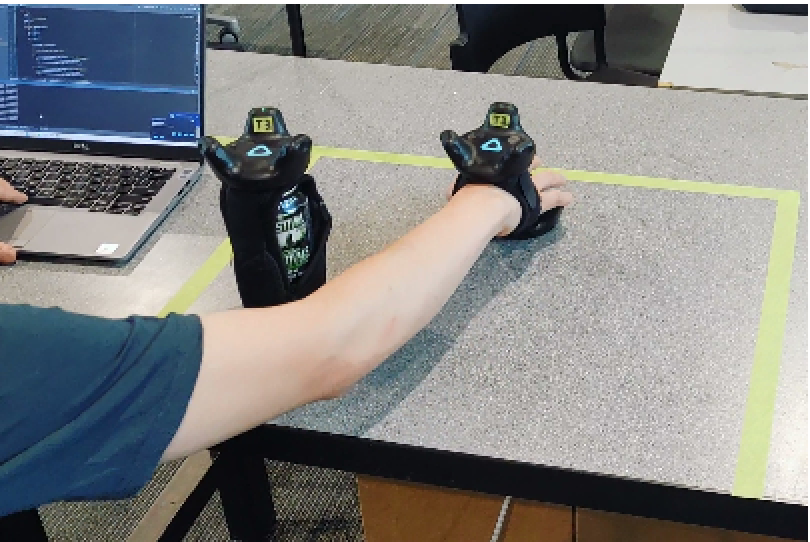}
        \label{exp_obs_4}}
    }
    \caption{The demonstration recording process. (a). Starting from the initial position. (b). Avoiding the obstacle. (c). Reaching the goal.}
    \label{fig:human_exp_photo}
\end{figure}

To develop the IBC-DMP agent, we have recorded a human demonstration dataset using this experimental setup.  As shown in Fig.~\ref{fig:human_exp_photo}, we use three trackers to mark the positions of the obstacle, the human hand, and the goal, respectively. We use a cylinder bottle with a height of $66\,$mm and a diameter of $200\,$mm as the obstacle. We fix the tracker to the top of the obstacle, leading to a total height of $70\,$mm. Besides, we tie a tracker to the wrist of the subject to track the motion of the hand. The third tracker is directly placed on the desk to mark the goal. During the entire recording process, the subject performs $600$ trials of P2PR motion repeatedly. For each trial, the goal $\mathcal{G}$ is placed at a random position in the motion region. It should be not too close to the initial position $\mathcal{I}$ to ensure the feasibility of the hand motion. The obstacle $\mathcal{O}$ is also positioned at a random point close to the mid-way between the initial position $\mathcal{I}$ and the goal $\mathcal{G}$. Note that the obstacle should be neither beside the goal nor too close to the initial position. Otherwise, it would be very difficult to produce a natural trajectory. 
The subject is asked to always perform natural and comfortable motions. Other than that, there are no restrictions from which direction the subject must avoid the obstacle. The subject is allowed to bend the torso but cannot stand up from the seat to reach a far point. 

\begin{figure}[htbp]
    \centering
    \subfloat[All hand trajectories]{\includegraphics[width=0.2\textwidth]{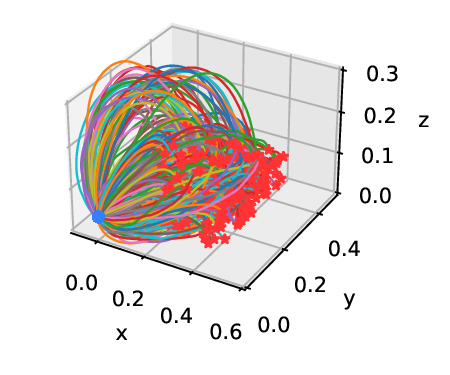}
        \label{fig:trials_2}}
    \subfloat[One hand trajectory]{\includegraphics[width=0.2\textwidth]{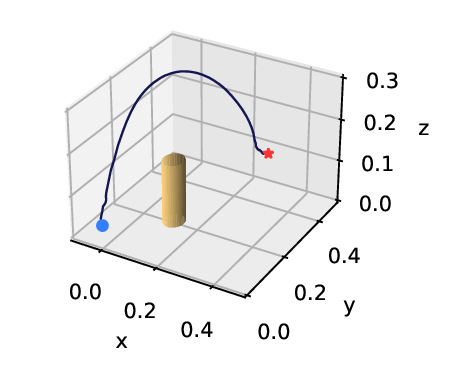}
        \label{fig:trials_4}}
    \caption{The recorded hand trajectories, where the blue dot denotes the starting position and the orange stars are the goal positions.}
    \label{fig:trials}
\end{figure}

After removing invalid records, we have obtained $544$ hand trajectories corresponding to various goal positions and obstacle positions, as illustrated in Fig.~\ref{fig:trials}. The figure clearly shows the various shapes of these trajectories brought up by human uncertainty. On the other hand, these trajectories present a common arc-like feature that implies smooth obstacle avoidance. These arc-shaped trajectories can be easily represented using motion primitives~\cite{paranjape2015motion} which provide a straightforward interface for robots to imitate. Nevertheless, the motion planning of a robot manipulator involves more constraints than simply imitating the shapes, including avoiding collisions with the ground and preventing singular configurations, making practical motion planning more challenging than pure imitation. 

\subsection{Data Preprocessing: Normalization}

The diversity of human hand trajectories is reflected by both their shapes and speeds. The shape of a trajectory prescribes in which direction the hand should move to avoid the obstacle, and its speed indicates how fast the hand moves. The diversity of the shapes is beneficial to improve the robustness of the trained planning agent against various obstacles and goal positions. However, the speeds of the hand trajectories do not contribute to the training robustness but bring up disturbances. Thus, a preprocessing procedure is needed to normalize the trajectory speeds while retaining the diversity of the shapes. In this paper, we are concerned with the average speed of a trajectory. Let $\pmb{\tau}=\{0,1, \cdots, \bar{T}\}$ be the time steps of a recorded hand trajectory $\mathbf{x}_{\tau}^H \in \mathbb{R}^3$, $\tau \in \pmb{\tau}$, at a discrete sampling time $\Delta \tau =0.01\,$s for the VIVE tracking system, where $\bar{T}$ is the number of samples per trajectory. Then, the average speed is defined as $\bar{V}=L/(\bar{T}\Delta \tau)$ (m/s), where $L=\sum_{\tau=1}^{\bar{T}} \|\mathbf{x}^H_{\tau} \!-\! \mathbf{x}^H_{\tau - 1}\|_2$ is the total length of the trajectory. 

We normalize the recorded trajectories to a uniform average speed $\bar{V}_*=0.2\,$m/s by resampling the trajectory at a new time interval $\displaystyle \Delta \tau' = {L}/{(\bar{V}_* \bar{T})}$, rendering new time steps $\pmb{\tau}'=\{0,1, \cdots, \bar{T}' \}$, where $\bar{T}' = \bar{T}\Delta \tau/\Delta \tau'$. Then, we approximate the linear velocity of the trajectory as $\displaystyle \dot{\mathbf{x}}^H_{\tau'} = {(\mathbf{x}^H_{\tau'}-\mathbf{x}^H_{\tau_{i}'- \Delta \tau'})}/{\Delta \tau'}$, $\tau'=\Delta \tau'$, $2\Delta \tau'$, $\cdots$, $\bar{T}'$, and calculate the maximal speed $\hat{V} = \max_i \|\dot{\mathbf{x}}^H_{\tau_i'}\|_2$. Similar to the average speed, the maximal speed is an important index to describe the kinematic property of a trajectory. We analyze the statistical properties of the total length, the maximal speed, and the average speed of all trajectories to investigate how the normalization changes their diversities. Their mean values (Mean), the standard deviations (Std), and the extreme deviations (Max-Min) are shown in Tab.~\ref{tab:p2p-ca}. The average speeds of the trajectories are normalized and the maximal speeds are reduced. Also, the deviation values of the maximal speeds are also reduced, which indicates less influence of the diversity of the maximal speeds on the recorded trajectories. Meanwhile, the statistical properties of the total lengths are not changed due to the consistency of the shapes of the trajectories. Therefore, the disturbance of the speeds of the trajectories is reduced while the diversity of the shapes is reserved.

\linespread{1.2}
\begin{table}[htbp]
\centering
\caption{The kinematic diversity of recorded trajectories}
\begin{tabular}{c|c|c|c|c|c|c}
	\thickhline
	\rowcolor{dark_red} \multirow{2}{*}{} & \multicolumn{3}{c|}{\color{white} \textbf{Before Normalization}} & \multicolumn{3}{c}{\color{white} \textbf{After Normalization}} \\ 
\cline{2-7}\rowcolor{mid_red} 
  & {\color{black} \textbf{Mean}} & {\color{black} \textbf{Std}} & {\color{black} \textbf{\scriptsize{Max-Min}}} & {\color{black} \textbf{Mean}} & {\color{black} \textbf{Std}} & {\color{black} \textbf{\scriptsize{Max-Min}}} \\
    \hline 
    $L$ & 0.4526 & 0.1779 & 1.1706 & 0.4526 & 0.1779 & 1.1706 \\
\rowcolor{shadow_red}    $\hat{V}$ & 1.1667 & 0.2787 & 1.6096 & 0.3998 & 0.0595 & 0.4807 \\
    $\bar{V}$ & 0.5793 & 0.1316 & 0.7842 & 0.2 & 0 & 0 \\
	\thickhline
\end{tabular}
\label{tab:p2p-ca}
\end{table}
\linespread{1}

\subsection{Translating Hand Trajectories Using DMP}\label{sec:data_gen}

After normalization, we generate human demonstration data from the processed hand trajectories. As implied by the overall framework shown in Fig.~\ref{fig:framework}, the human demonstration data to be generated must fit the same multi-DoF DMP used to train the planning agent, for which the translation of the normalized hand trajectories is needed. This ensures that any robot platform with a general position-based control interface can reproduce the recorded hand trajectories.

To ensure the consistent sampling rate with the discrete-time DMP model in Eq. \eqref{eq:dmp_multi_dof_dis}, we resample the normalized trajectory $\x^H_{\tau'}$, $\tau \in \pmb{\tau}'$, using $\Delta t$, denoted as $\x^H_{t}$, $t \in \{0,1,\cdots,T\}$. As addressed in Sec.~\ref{sec:mdp}, the data for the experience replay buffer have the format $\{\,s_t, a_t, r_t, s'_t, d_t\,\}$, for a certain time $t\in \mathbb{Z}^+$, where the state $s_t = \mathcal{X}_t$ is also the state of the multi-DoF DMP model as introduced in Sec.~\ref{sec:multi_dmp}. The action $a_t \!=\! \tilde{\mathbf{f}}\!\left(\mathcal{X}_t \right)$ is the output of the demonstration policy $\tilde{\mathbf{f}}$ which allows $\x_t$ to fit the multi-DoF DMP model. Then, $s'\!=\!\mathcal{X}_{t+1}$ is the successive state of $s_t$ under the action $a_t$, $r_t\!=\!-\mathcal{J}_t$ is the instant reward, and $d_t$ is a binary value to determine whether $\x_t$ reaches $\mathbf{x}_{\mathrm{g}}$. 

The determination of action $a_t$ is not trivial since the demonstration policy $\tilde{\mathbf{f}}$ is not previously known. In conventional work, each action sample $a_t=\tilde{\mathbf{f}}(\mathcal{X}_t)$ for a given state sample $\mathcal{X}_t$ is directly computed by inverting the DMP model~\eqref{eq:dmp_multi_dof}. This requires calculating the acceleration $\ddot{\x}_t$ using a twice-difference operation which may bring differential noise to the action samples. As a result, the variance of the samples may be increased, which may disturb the training process. In this paper, we use a PID-based approach to regenerate actions for the state samples. Specifically, by representing the positions and velocities of a human hand trajectory as $\x_t^H, \dot{\x}_t^H \!\in\! \mathbb{R}^3$, we use a multi-DoF DMP model described by \eqref{eq:dmp_multi_dof} and \eqref{eq:dmp_multi_dof_dis} with a virtual force function $\tilde{\mathbf{f}}(\mathcal{X}_t) = K_{\mathrm{P}} (\x_t^H\!-\!\x_t) + K_{\mathrm{D}} (\dot{\x}_t^H\!-\!\dot{\x}_t)$ to generate a trajectory $\x_t$ that fits the hand trajectory $\x^H_t$, with initial and goal conditions $\x_{0}\!=\!\x_{0}^H$, $\x_{\mathrm{g}}\!=\!\x_{\mathrm{g}}^H$, and $\dot{\x}_{0}\!=\!0$, where $K_{\mathrm{P}}=1500 I$ and $K_{\mathrm{P}} =40 I$ are constant matrices, and $I\!\in\! \mathbb{R}^{3\times 3}$ is an identity matrix. The main advantage of the PID-based method is not requiring the acceleration $\ddot{\x}_t$. Thus, it can reduce the noise introduced to the samples. With proper parameters $K_{\mathrm{P}}$, $K_{\mathrm{P}}$, the generated trajectory $\x_t$ coincides with the recorded hand trajectory $\ddot{\x}_t$ with small errors $\x_t^H - \x_t$, which ensures the validity of the action samples.

Having determined the action sample $a_t$ for the state sample $s_t$, the successive state $s'_t$ can also be calculated using the multi-DoF DMP model. The reward $r_t$ can be calculated using $-\mathcal{J}_t$ from the reward function \eqref{eq:cost}. Then, the termination flag $d_t$ is determined as follows,
\begin{equation}\label{eq:term_flag}
d_t = \left\{ \begin{array}{ll}
1 & \mathrm{if}\, t = T\, \mathrm{or} \, \|\x_t - \x_{\mathrm{g}}\|_2 \leq \varepsilon_T, \\
0 & \mathrm{otherwise},
\end{array}\right.
\end{equation}
where $T$ is the maximal length of an episode.

We ultimately transform 544 recorded hand trajectories into 123171 samples. 
The parameters of the multi-DoF DMP model and the cost function $\mathcal{J}_t$ are shown in Tab.~\ref{tab:hyper-param-dmp}. 

\linespread{1.2}
\begin{table}[htbp]
\centering
\caption{Parameters of Multi-DoF DMP and costs}
\begin{tabular}{c|c||c|c||c|c||c|c}
	\thickhline
	\rowcolor{dark_blue} {\color{white} \textbf{Par.}} & {\color{white} \textbf{Val.}} & {\color{white} \textbf{Par.}} & {\color{white} \textbf{Val.}} & {\color{white} \textbf{Par.}} & {\color{white} \textbf{Val.}} & {\color{white} \textbf{Par.}} & {\color{white} \textbf{Val.}} \\
        \hline
        $K_{\alpha}$ & $10 I$ & $K_{\beta}$ & $1.2 I$ & $\tau$ & $0.25$ & $\omega$ & $6$ \\
        \rowcolor{shadow_blue} $\underline{f}$ & $-5$ & $\overline{f}$ & $5$ & $\alpha_1$ & $0.001$ & $\alpha_2$ & $10$ \\
        $\alpha_3$ & $0.001$ & $\alpha_4$ & $0.001$ & $\alpha_5$ & $10^5$ & $\varepsilon_T$ & $0.01$ \\
        \rowcolor{shadow_blue} $\varepsilon_0^{\mathrm{b}}$ & $0.05$ & $\varepsilon_1^{\mathrm{b}}$ & $0.045$ & $\varepsilon_0^{\mathrm{d}}$ & $0.01$ & $\varepsilon_1^{\mathrm{d}}$ & $0.05$ \\
	\thickhline
\end{tabular}
\label{tab:hyper-param-dmp}
\end{table}
\linespread{1}

\section{Training of an IBC-DMP agent}\label{sec:training}

This section interprets the training of an IBC-DMP agent. We first give an overview of the training procedure. Then, we specifically interpret its critical technical points. Finally, we formulate the training algorithm for an IBC-DMP agent.

\subsection{The Training Overview}\label{sec:m_overview}

\begin{figure}[htbp]
\noindent
\hspace*{\fill} 
\begin{tikzpicture}[scale=1,font=\small]

\def\sbheight{0.8cm}
\def\sbwidth{1.6cm}
\def\dev{1.5cm}

\definecolor{closedloop}{RGB}{0, 0, 255}

\definecolor{darkgray}{RGB}{220, 220, 220}
\definecolor{shadowgray}{RGB}{245, 245, 245}
\definecolor{shadowred}{RGB}{255, 229, 229}
\definecolor{shadowyellow}{RGB}{255, 255, 229}
\definecolor{shadowblue}{RGB}{229, 240, 255}
\definecolor{shadowgreen}{RGB}{245, 255, 245}
\definecolor{shadoworange}{RGB}{255, 245, 229}

\node[minimum height=2.7cm, minimum width=3.2cm, draw, thick, fill=shadoworange, rounded corners=0.3cm] (demo) at (0cm,0.9cm) {};
\node[anchor=north west] at (demo.north west) {\scriptsize  ~~\textbf{Demostration Buffer}};
\node[minimum height=0.5cm, inner sep=0pt, minimum width=3cm, draw, thick, fill=shadowyellow, rounded corners=0.25cm] (item_1) at (0cm, 1.6cm) {\scriptsize $\iota_1^{\mathrm{D}} \!=\! \left\{s_t, a_t, r_t, s_t', d_t \right\}$};
\node[minimum height=0.5cm, inner sep=0pt, minimum width=3cm, draw, thick, fill=shadowyellow, rounded corners=0.25cm] (item_2) at (0cm, 1.0cm) {\scriptsize $\iota_2^{\mathrm{D}} \!=\!  \left\{s_t, a_t, r_t, s_t', d_t \right\}$};
\node[] at (0cm, 0.5cm) {$\cdots$};
\node[minimum height=0.5cm, inner sep=0pt, minimum width=3cm, draw, thick, fill=shadowyellow, rounded corners=0.25cm] (item_3) at (0cm, 0cm) {\scriptsize $\iota_n^{\mathrm{D}} \!=\! \left\{s_t, a_t, r_t, s_t', d_t \right\}$};

\node[minimum height=5cm, minimum width=3.2cm, draw, thick, fill=shadoworange, rounded corners=0.3cm] (replay) at (4cm, 2.2cm) {};
\node[anchor=north west] at (replay.north west) {\scriptsize  ~~\textbf{Replay Buffer}};
\node[minimum height=0.5cm, inner sep=0pt, minimum width=3cm, draw, thick, fill=shadowyellow, rounded corners=0.25cm] (item_4) at (4cm, 4cm) {\scriptsize  $\iota_1^{\mathrm{D}} \!=\! \left\{ s_t, a_t, r_t, s_t', d_t \right\}$};
\node[minimum height=0.5cm, inner sep=0pt, minimum width=3cm, draw, thick, fill=shadowyellow, rounded corners=0.25cm] (item_5) at (4cm, 3.4cm) {\scriptsize  $\iota_2^{\mathrm{D}} \!=\! \left\{ s_t, a_t, r_t, s_t', d_t \right\}$};
\node[] at (4cm, 2.9cm) {$\cdots$};
\node[minimum height=0.5cm, inner sep=0pt, minimum width=3cm, draw, thick, fill=shadowyellow, rounded corners=0.25cm] (item_6) at (4cm, 2.4cm) {\scriptsize  $\iota_n^{\mathrm{D}} \!=\! \left\{ s_t, a_t, r_t, s_t', d_t \right\}$};
\node[minimum height=0.5cm, minimum width=3cm, draw, thick, fill=shadowyellow, rounded corners=0.25cm] (item_7) at (4cm, 1.7cm) {\scriptsize  $\iota_1 \!=\! \left\{ s_t, a_t, r_t, s_t', d_t \right\}$};
\node[minimum height=0.5cm, minimum width=3cm, draw, thick, fill=shadowyellow, rounded corners=0.25cm] (item_8) at (4cm, 1.1cm) {\scriptsize  $\iota_2 \!=\! \left\{ s_t, a_t, r_t, s_t', d_t \right\}$};
\node[] at (4cm, 0.6cm) {$\cdots$};
\node[minimum height=0.5cm, minimum width=3cm, draw, thick, fill=shadowyellow, rounded corners=0.25cm] (item_9) at (4cm, 0.1cm) {\scriptsize  $\iota_k \!=\! \left\{ s_t, a_t, r_t, s_t', d_t \right\}$};

\node[minimum height=0.6cm, minimum width=1.2cm, draw, thick, fill=shadowgreen, rounded corners=0.1cm] (bc_loss) at (-0.7cm,-2.3cm) {\scriptsize \textbf{BC Loss}};
\node[circle,inner sep=0pt,minimum size=0.25cm,draw,fill=darkgray] (plus1) at (0.8cm, -2.3cm) {};
\draw[thick] (plus1.north east) -- (plus1.south west);
\draw[thick] (plus1.north west) -- (plus1.south east);
\node[anchor=south west] at ([xshift=-0.1cm, yshift=-0.1cm] plus1.north east) {\scriptsize $+$};
\node[minimum height=0.6cm, minimum width=1.5cm, draw, thick, fill=shadowgreen, rounded corners=0.1cm] (pi_loss) at (2cm,-2.3cm) {\scriptsize \textbf{Policy Loss}};
\node[minimum height=0.6cm, minimum width=1.5cm, draw, thick, fill=shadowgreen, rounded corners=0.1cm] (val_loss) at (4cm,-2.3cm) {\scriptsize \textbf{Value Loss}};
\draw[->,>=stealth,thick,densely dashed,double] ([xshift=-0.7cm] demo.south) -- node[pos=0.8, anchor=west]{$\mathcal{B}_{\mathrm{D}}^{\pi}$} (bc_loss.north);
\draw[->,>=stealth,thick,densely dashed,double] (replay.west) -- ([xshift=-0.4cm] replay.west) -- ([yshift=0.5cm] pi_loss.north) -- node[pos=0.8, anchor=south west] {$\mathcal{B}_{\mathrm{R}}^{\pi}$} (pi_loss.north);
\node[anchor=south west] () at (val_loss.north) {$\mathcal{B}^{Q}$};

\node[minimum height=0.1cm, inner sep=0pt, minimum width=0.8cm, fill=black] (joint) at ([yshift=0.5cm] val_loss.north) {};
\draw[->,>=stealth,thick,densely dashed,double] (joint.south) -- (val_loss.north);
\draw[->,>=stealth,thick,densely dashed,double] ([xshift=0.2cm] replay.south) -- node[pos=1, anchor=south west]{$\mathcal{B}_{\mathrm{R}}^{Q}$} ([xshift=0.2cm] joint.north);
\draw[->,>=stealth,thick, thick] (bc_loss.east) -- node[pos=0.5, anchor=south]{\scriptsize $\times \lambda_{\mathrm{BC}}$} (plus1.west);
\draw[->,>=stealth,thick, thick] (pi_loss.west) -- (plus1.east);

\node[minimum height=5cm, minimum width=5.5cm, draw, thick, dotted, fill=shadowgray, rounded corners=0.4cm] (agent) at (2.7cm,-5.9cm) {};
\node[minimum height=2.4cm, minimum width=1.6cm, draw, thick, fill=shadowred, rounded corners=0.3cm, rotate=-90] (actor) at ([xshift=-1.3cm, yshift=1.4cm] agent.center) {\includegraphics[width=1.5cm]{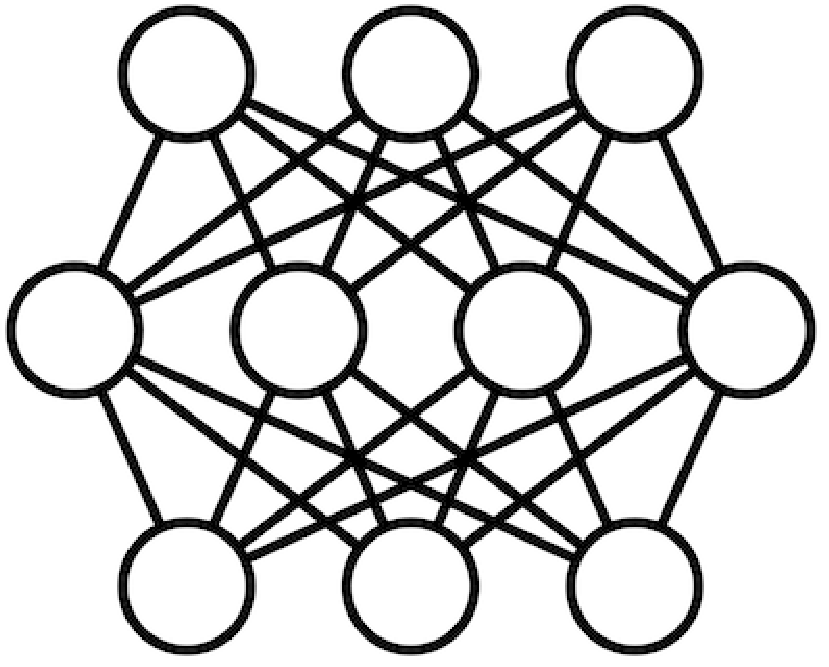}};
\node[minimum height=2.4cm, minimum width=1.6cm, draw, thick, fill=shadowred, rounded corners=0.3cm, rotate=-90] (critic) at ([xshift=1.3cm, yshift=1.4cm] agent.center) {\includegraphics[width=1.5cm]{nn_crop.eps}};
\node[anchor=north west] at (actor.south west) {\scriptsize Actor};
\node[anchor=north west] at (critic.south west) {\scriptsize Critic};
\node[anchor=north west] at (actor.south east) {\scriptsize ~~~~~~~~~~~~~~~~~\textbf{Source Networks}};
\node[anchor=south west] at (agent.north west) {\textbf{Agent}};

\node[minimum height=2.4cm, minimum width=1.6cm, draw, thick, fill=shadowblue, rounded corners=0.3cm, rotate=-90] (actor_tar) at ([xshift=-1.3cm, yshift=-1.1cm] agent.center) {\includegraphics[width=1.5cm]{nn_crop.eps}};
\node[minimum height=2.4cm, minimum width=1.6cm, draw, thick, fill=shadowblue, rounded corners=0.3cm, rotate=-90] (critic_tar) at ([xshift=1.3cm, yshift=-1.1cm] agent.center) {\includegraphics[width=1.5cm]{nn_crop.eps}};
\node[anchor=north west] at (actor_tar.south west) {\scriptsize Actor};
\node[anchor=north west] at (critic_tar.south west) {\scriptsize Critic};
\node[anchor=north west] at (actor_tar.south east) {\scriptsize ~~~~~~~~~~~~~~~~~\textbf{Target Networks}};

\draw[->,>=stealth,thick, thick, double] (plus1.south) -- ([yshift=-0.4cm] plus1.south) -- ([xshift=0.6cm, yshift=-0.4cm] plus1.south) -- node[pos=0.4, anchor=west] {$\mathcal{L}_A$} (actor.west);
\draw[->,>=stealth,thick, thick, double] (val_loss.south) -- ([yshift=-0.3cm]val_loss.south) -- ([yshift=0.7cm] critic.west) -- node[pos=0.3, anchor=east] {$\mathcal{L}_C$} (critic.west);

\draw[->,>=stealth,thick,densely dashed,double] ([xshift=0.7cm] demo.south) -- ([xshift=0.7cm, yshift=-0.4cm] demo.south) -- ([xshift=-0.2cm, yshift=0.6cm] joint.north) -- node[pos=0.5, anchor=east]{$\mathcal{B}_{\mathrm{D}}^{Q}$} ([xshift=-0.2cm] joint.north);

\node[minimum height=1.4cm, minimum width=2cm, draw, thick, fill=white, rounded corners=0.1cm] (demo_traj_0) at (0.2cm, 3.9cm) {};
\node[minimum height=1.4cm, minimum width=2cm, draw, thick, fill=white, rounded corners=0.1cm] (demo_traj_1) at (0.1cm, 3.8cm) {};
\node[minimum height=1.4cm, minimum width=2cm, draw, thick, fill=white, rounded corners=0.1cm] (demo_traj_2) at (0cm, 3.7cm) {};
\node[minimum height=1.4cm, minimum width=2cm, draw, thick, fill=white, rounded corners=0.1cm] (demo_traj_3) at (-0.1cm, 3.6cm) {\includegraphics[width=1.4cm]{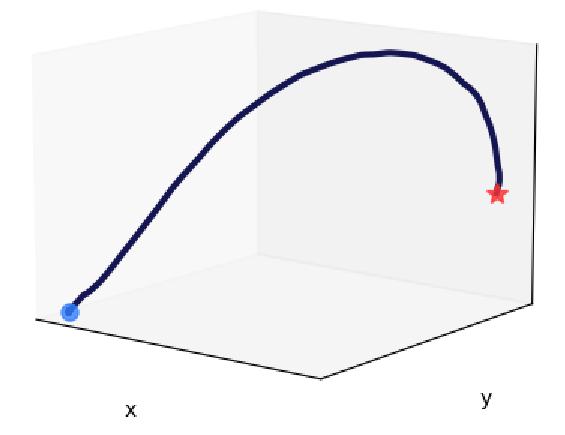}};
\node[anchor=north west] at (demo_traj_3.north west) {\scriptsize \textbf{Trajectories}};

\draw[->,>=stealth, thick] (demo_traj_3.south) -- node[pos=0.3, anchor=west]{$\iota^{\mathrm{D}}$} ([xshift=-0.1cm] demo.north);
\draw[->,>=stealth, thick] (demo_traj_3.east) -- node[pos=0.5, anchor=north] {$\iota^{\mathrm{D}}$} ([yshift=1.4cm] replay.west);

\draw[->,>=stealth, thick,double] (actor.east) -- node[pos=0.5, anchor=east]{$\theta$} (actor_tar.west);
\draw[->,>=stealth, thick,double] (critic.east) -- node[pos=0.5, anchor=west]{$w$} (critic_tar.west);

\node[circle,inner sep=0pt,minimum size=4mm,draw,color=black] (swr) at (-1.5cm, 3.5cm) {1};
\node[circle,inner sep=0pt,minimum size=4mm,draw,color=black] (swr) at (-1.5cm, -1cm) {2};
\node[circle,inner sep=0pt,minimum size=4mm,draw,color=black] (swr) at (-0.5cm, -5.6cm) {3};
\node[circle,inner sep=0pt,minimum size=4mm,draw,color=black] (swr) at (5.6cm, -1cm) {4};

\node[circle,inner sep=0pt,minimum size=7mm,draw, fill=shadowyellow] (iotak1) at ([xshift=1.2cm,  yshift=-1.5cm] replay.south) {$\iota_{k+1}$};
\draw[<-,>=stealth, thick, dotted] ([xshift=1.2cm] replay.south) -- (iotak1.north);

\node[minimum height=5.1cm, minimum width=3.5cm, color=red, draw, very thick, dashed] (ibc) at ([yshift=-1.1cm] demo.center) {};
\node[anchor=south west] () at (ibc.north west) {\color{red} \textbf{IBC}};

\node[minimum height=2.3cm, minimum width=3.5cm, color=blue, draw, very thick, dashed] (dir) at ([yshift=1.0cm] replay.center) {};
\node[anchor=north east] () at (dir.north west) {\color{blue} \textbf{DIR}};

\end{tikzpicture}
\hspace{\fill} 
\caption{The flow chart of training an IBC-DMP agent, with circled numbers denoting critical steps to be explained later.}
\label{fig:over_view}
\end{figure}

The training process of an IBC-DMP agent is illustrated as a flow chart in Fig.~\ref{fig:over_view}. Similar to a traditional DDPG agent, the IBC-DMP agent consists of four forward neural networks (FNN), namely source and target actors $\pi_{\theta},\pi_{\theta'}$ and source and target critics $Q_w,Q_{w'}$, where the subscripts $\theta$, $\theta'$, $w$ and $w'$ denote the parameters of these networks. The source actor and critic are constructed to approximate the optimal policy and the value function of the RL problem, respectively. Meanwhile, the target networks are used to improve the stability of this approximation.

Different from a conventional DDPG agent, an IBC-DMP agent has two data buffers, namely a \textit{demo buffer} and a \textit{replay buffer}, inspired by the previous work on dual-buffer off-policy RL~\cite{ying2022trajectory, gupta2022exploiting}. The \textit{demo buffer} (marked as a red dashed box in Fig.~\ref{fig:over_view}) stores the demonstration data of the human motion recorded in Sec.~\ref{sec:human_data}, used to promote the learning of the agent via the \textit{IBC} technology. Meanwhile, the \textit{replay buffer} serves as a normal experience replay storage unit but is initialized using the demonstration data, which is different from a conventional DDPG approach. We refer to this initialization process as \textit{demonstration initialization for replay (DIR)} and highlight it using a blue dashed box in Fig.~\ref{fig:over_view}.

The DIR technology originated from a very straightforward idea of utilizing demonstration for RL, namely mixing the demonstration data and the experience data in the replay buffer. Nevertheless, these two data sources encode diverse policies and may lead to unexpected extrapolations, potentially causing overfitting, as explained in Sec.~\ref{sec:bc}. To mitigate this effect, DIR only imports demonstration to the replay buffer during initialization, allowing the demonstration to be forgotten when the agent is well trained. Based on this, IBC further promotes the learning speed by imitating the high-value actions from the demonstration, which is to be explained in the next subsection.

\subsection{The Training Details}

The training steps for IBC-DMP are summarized as follows.

\subsubsection{Demonstration Data Importing}

The human demonstration data generated in Sec.~\ref{sec:human_data} are imported to the demo buffer and then shuffled. They are also imported to the replay buffer to boot up the training process. The importing processes are denoted as solid arrows in Fig.~\ref{fig:over_view}. This is executed only once at the beginning of each training process.

\subsubsection{Batch Sampling}

During the training process, four data batches are regularly sampled from the two buffers, denoted as $\mathcal{B}_{\mathrm{D}}^{\pi}$, $\mathcal{B}_{\mathrm{D}}^{Q}$, $\mathcal{B}_{\mathrm{R}}^{\pi}$, and $\mathcal{B}_{\mathrm{R}}^{Q}$, respectively, where the subscripts $\mathrm{D}$ and $\mathrm{R}$ and the superscripts $\pi$ and $Q$ indicate from the demo buffer, from the replay buffer, for the actor, and for the critic, respectively. The sampling is conducted individually and independently to eliminate the dependence between the samples. The sampling process is represented as double dashed arrows in Fig.~\ref{fig:over_view}. The sampling frequency of the data batches is the same as the update rate of the networks.

\subsubsection{Network Updates}

The sampled data batches are used to calculate the losses of the actor and the critic using loss functions $\mathcal{L}_A$ and $\mathcal{L}_C$, respectively. In this paper, we propose novel loss functions for the training of the neural networks of the IBC-DMP agent, which will be introduced in Sec.~\ref{sec:bc}. 
With the loss functions $\mathcal{L}_A$ and $\mathcal{L}_C$, the parameters of the source networks are updated using the gradient-based law in \eqref{eq:ac_grad}. Then, the target networks are updated using \eqref{eq:tac_grad}. This process is denoted by double arrows in Fig.~\ref{fig:over_view}.

\subsubsection{Experience Storing and Forgetting}

Similar to all off-policy RL agents with experience replay, the IBC-DMP agent also stores its interaction data with the environment in the replay buffer, at every sampling instant. As shown in Fig.~\ref{fig:over_view}, the latest interaction data is always added to the tail of the replay buffer, right after the demonstration data. The storing process is represented as a dashed arrow. The replay buffer has a fixed size such that the old data is forgotten and replaced by the new data. Thus, the demonstration data are ultimately purged from the replay buffer and lose their impacts on the data batches $\mathcal{B}_{\mathrm{D}}^Q$ and $\mathcal{B}_{\mathrm{R}}^{Q}$ as the learning proceeds. 

\subsection{Reshaped Actor Loss Based on IBC}\label{sec:ibc}

One critical technical point of IBC-DMP is to properly design the loss functions for the networks. We propose a reshaped loss function for the actor,
\begin{equation}\label{eq:la}
\mathcal{L}_A \!\left(\mathcal{B}_{\mathrm{D}}^{\pi}, \mathcal{B}_{\mathrm{R}}^{\pi} \right) = \hat{\mathcal{L}}_{A}\!\left(\mathcal{B}_{\mathrm{R}}^{\pi} \right) + \lambda_{\mathrm{BC}} \mathcal{L}_{BC} \!\left(\mathcal{B}_{\mathrm{D}}^{\pi} \right),
\end{equation}
where $\hat{\mathcal{L}}_{A}\!\left(\mathcal{B}_{\mathrm{R}}^{\pi} \right)$ has the same form as a conventional actor loss function defined in \eqref{eq:conv_actor_loss}, $\mathcal{L}_{BC}\!\left(\mathcal{B}_{\mathrm{D}}^{\pi} \right)$ is a novel IBC loss function and $\lambda_{\mathrm{BC}} \in \mathbb{R}^+$ is the BC ratio to adjust the proportion of $\lambda_{\mathrm{BC}}$ in the overall actor loss $\mathcal{L}_A$. For any data buffer $\mathcal{B}$ sized $n \in \mathbb{Z}^+$, the IBC loss is calculated as
\begin{equation}\label{eq:bc_loss}
\textstyle \mathcal{L}_{BC}\!\left(\mathcal{B} \right) \!=\! -\frac{1}{n} \!\sum_{j=1}^{n} E_w(\pi_{\theta}|s_j,a_j),
\end{equation}
where $E_w(\pi_{\theta}|s_j,a_j)$ is an energy function of $\pi_{\theta}$ defined as
\begin{equation}\label{eq:energf}
E_w(\pi_{\theta}|s_j,a_j) \!=\! -\mathrm{ReLU}\! \left( Q_w(s_j, a_j)\!-\! Q_w(s_j, \pi_{\theta}(s_j)) \right),
\end{equation}
where $\{s_j, a_j\}$ are the $j$-th state and action samples of $\mathcal{B}$, $j = 0, 1, \cdots, n$, and $\mathrm{ReLU}(x) = \max(x,0)$, $x\in \mathbb{R}$, is a Rectified Linear Unit (ReLU) function.

The actor loss function in \eqref{eq:la} consists of two parts. The first part $\hat{\mathcal{L}}_A$ is defined on the interaction data batch $\mathcal{B}_{\mathrm{R}}^{\pi}$ and has the same form as the conventional actor loss function of an off-policy RL agent. The second part $\mathcal{L}_{BC}$ is a novel IBC-based loss function defined on the demonstration data batch defined on the interaction data batch $\mathcal{B}_{\mathrm{D}}^{\pi}$. It is used to penalize the current policy $\pi_{\theta}$ if produces a worse action $\pi_{\theta}(s_j)$ than the demonstration action $a_j$. As a result, it forces the policy $\pi_{\theta}$ to perform better than the demonstration policy during the training process. The extent of this effect is controlled by the BC ratio $\lambda_{\mathrm{BC}}$. In this sense, BC is seamlessly integrated into the training process of the actor, which leads to a flexible manner of updating the policies.

The IBC loss function defined in Eq.~\eqref{eq:bc_loss} is based on the energy function given in Eq.~\eqref{eq:energf} which penalizes the policy $\pi_{\theta}$ when it performs worse (in the sense of approximated value) than the demonstration. Different from the EBC-based RL in previous work which forces the current policy $\pi_{\theta}$ to imitate the mixed demonstration policy and the target policy~\cite{gupta2021reinforcement}, it ensures that the policy $\pi_{\theta}$ only fits the superior one between the target policy and the demonstration policy, instead of both. This can effectively avoid unexpected extrapolations between these two policies~\cite{florence2022implicit}, thus resolving the overfitting issue. 


\subsection{Refined Critic Loss}

The IBC loss in \eqref{eq:conv_actor_loss} depends on the parameter $w$ of the critic $Q_w$. If $Q_w$ is not well trained, the BC loss $\mathcal{L}_{BC}$ may not be able to 
accurately capture the penalty that should be imposed on the current policy $\pi_{\theta}$, leading to an invalid loss that does not help the policy update. This is especially likely to occur in the initial stage of the learning process. 
To solve this problem, we use a refined data batch $\mathcal{B}^{Q} \!=\! \mathcal{B}_{\mathrm{D}}^{Q} \cup \mathcal{B}_{\mathrm{R}}^{Q}$ to compose the loss function for the critic, $\mathcal{L}_C\!\left(\mathcal{B}^{Q} \right)$, where the form of $\mathcal{L}_C$ is defined in \eqref{eq:conv_critic_loss}.
Here, we refer to $\mathcal{B}^{Q}$ as a refined data batch due to a larger proportion of demonstration data compared to a replay batch $\mathcal{B}_{\mathrm{R}}^{Q}$ of the same size. The ratio between the sizes of $\mathcal{B}_{\mathrm{D}}^{Q}$ and $\mathcal{B}_{\mathrm{R}}^{Q}$ is referred to as the \textbf{refining factor} $\lambda_{\mathrm{RF}}$. Here, we assume that human demonstration data likely have higher values than random data. Therefore, a refined data batch with a larger $\lambda_{\mathrm{RF}}$ is likely to better describe the true critic loss and thus is more helpful to the booting up of the training of the critic. However, an overlarge $\lambda_{\mathrm{RF}}$ may lead to overfitting to the human demonstration.
In this paper, we select $\lambda_{\mathrm{RF}}$ as a constant.
For better performance, the demo batch $\mathcal{B}_{\mathrm{D}}^{Q}$ can be gradually eliminated from $\mathcal{B}^{Q}$ as the learning proceeds, which renders a decreasing refining factor $\lambda_{\mathrm{RF}}$. This can be an interesting topic for future work.

\subsection{The IBC-DMP Training Algorithm}

The training procedure of an IBC-DMP agent is shown in Algorithm \ref{ag:demo}, where $N$ is the total number of episodes during the training process, $T$ is the maximal number of steps in an episode, and $\epsilon_t \sim \mathcal{N}(0, \sigma^2)$ is a random variable for a certain time $t = 0, 1, \cdots, T$. Algorithm \ref{ag:demo} inherits the following common points from the conventional off-policy RL method.
\begin{itemize}[leftmargin=*]
\item Initialization of the four neural networks (lines 2 and 3).
\item Random sampling of actions (line 9).
\item Storing interaction data (line 13).
\item Sampling data from the replay buffer (line 16).
\item Updating networks (lines 19 and 20).
\end{itemize}

Meanwhile, Algorithm~\ref{ag:demo} is different from the conventional off-policy RL agent in the following aspects.
\begin{itemize}[leftmargin=*]
\item Initialization of the \textit{demo} and \textit{replay} buffers (line 1).
\item A multi-DoF DMP as the environmental model (line 10).
\item Sampling data from the demo buffer (line 15).
\item The computation of loss functions (lines 17 and 18).
\end{itemize}

\begin{algorithm}[htbp]
\caption{The training of an IBC-DMP agent}\label{ag:demo}
\begin{algorithmic}[1]
\renewcommand{\algorithmicrequire}{\textbf{Input:}}
\renewcommand{\algorithmicensure}{\textbf{Output:}}
\REQUIRE Demo buffer and replay buffer
\ENSURE Trained policy parameter $\theta$
\STATE Import demonstration data $\iota^{\mathrm{D}}$ to both buffers
\STATE Randomly initialize the source networks $\pi_{\theta}$, $Q_{w}$
\STATE Initialize the target networks $\pi_{\theta'}$, $Q_{w'}$ with $\theta' \!\leftarrow\! \theta$, $w' \!\leftarrow\! w$

\FOR{$i \leftarrow 1$ \TO $N$}
    \STATE Sample the initial position $\x_{\mathrm{i}}$, the obstacle position $\x_{\mathrm{b}}$, and the goal position $\x_{\mathrm{g}}$
    \STATE Initialize the state $s_0 \!=\! \mathcal{X}_{0}$
    \FOR{$t \leftarrow 0$ \TO $T$}
        \STATE Observe the state $s_t \!=\! \mathcal{X}_{t}$
        \STATE Sample an action $a_t=\pi_{\theta}(s_t) + \epsilon_t$
        \STATE Update the DMP model using \eqref{eq:dmp_multi_dof_dis}
        \STATE Observe the successive state $s_t'\!=\!\mathcal{X}_{t+1}$
        \STATE Calculate the instant reward $r_t$ and flag $d_t$
        \STATE Store $\{s_t, a_t, s_t', r_t, d_t\}$ to the replay buffer
        \IF{\textit{UPDATE} is \textbf{true}}
        \STATE Sample batches $\mathcal{B}_{\mathrm{D}}^{\pi}$, $\mathcal{B}_{\mathrm{D}}^Q$ from demo buffer
        \STATE Sample batches $\mathcal{B}_{\mathrm{R}}^{\pi}$, $\mathcal{B}_{\mathrm{R}}^{Q}$ from replay buffer
        \STATE Calculate the actor loss $\mathcal{L}_A\!\left(\mathcal{B}_{\mathrm{D}}^{\pi}, \mathcal{B}_{\mathrm{R}}^{\pi} \right)$ using \eqref{eq:la}
        \STATE Calculate the critic loss $\mathcal{L}_B\!\left(\mathcal{B}_{\mathrm{D}}^{Q} \!\cup\! \mathcal{B}_{\mathrm{R}}^{Q} \right)$ using \eqref{eq:conv_critic_loss}
        \STATE Update the source networks $\pi_{\theta}$, $Q_w$ using~\eqref{eq:ac_grad}
        \STATE Update the target networks $\pi_{\theta'}$, $Q_{w'}$ using~\eqref{eq:tac_grad}
        \ENDIF
    \STATE $s_t \leftarrow s_t'$
    \ENDFOR

\ENDFOR

\end{algorithmic}
\end{algorithm}

\section{Experiments for Evaluation}\label{sec:va_sim}

This section showcases the efficacy of the proposed IBC-DMP RL method using comparison and ablation studies.
We first introduce the setting of the experiment, followed by the detailed configuration of the IBC-DMP agent and its ablated agents. Then, we investigate the training and test performance of the agents, respectively.
The code of the studies is implemented based on the Spinningup baseline programs~\cite{SpinningUp2018} and written in Python. The training and test are performed on a Thinkpad laptop workstation with Intel(R) Core(TM) i7-10750H CPU at $2.60\,$GHz.
The programs of this study are published at 
https://doi.org/10.5281/zenodo.11237604.

\subsection{Experimental Configuration}\label{sec:domains}

We consider a P2PR task similar to the human trajectory recording experiment in Sec.~\ref{sec:human_data}. Specifically, the end-effector of a robot is expected to move from a fixed initial position $\x_{\mathrm{i}}\!=\!(0,\,0,\,0.05)\,$m to a random goal position $\x_{\mathrm{g}}$ while avoiding collisions with a cylinder obstacle placed at a random position $\x_{\mathrm{b}}$. Both positions are sampled from a $0.6\,$m long, $0.5\,$m wide, and $0.1\,$m height rectangular experimental domain $\mathcal{D}$ centered at $(0.32~0.34~0.09)\,$m, covering most of the reachable space of the robot in the first quadrant of the horizontal Cartesian coordinate. Most goal positions in the human data recording experiment are also in this region. Thus, this domain is sufficiently representative to reflect the true generalizability of a motion planning agent.
The radius and height of the obstacle are $r_{\mathrm{b}}=0.035\,$m and $\x_{\mathrm{b}}^{(3)} = 0.16\,$m, the same as the demonstration recording experiment in Sec.~\ref{sec:human_data}. 

Such a configuration allows a simplified and straightforward experimental setting while incorporating typical and common challenges in robot motion planning. Firstly, assigning a fixed initial position $\x_{\mathrm{i}}$ does not lose the generality since we can always use a coordinate transformation to transform any initial position in practice to $\x_{\mathrm{i}}$. Secondly, we treat the robot end-effector as a point and only consider its position, which does not eliminate the true challenge of motion planning. Thirdly, the experimental configuration is similar to the human trajectory recording experiment in Sec.~\ref{sec:human_data}. This likely generates similar arc-shaped trajectories for the robot. Nevertheless, as mentioned in Sec.~\ref{sec:hdr}, robot motion planning needs to incorporate additional constraints than purely imitating arc-shaped trajectories, including avoiding collisions with the ground and preventing singular configurations. In this sense, this configuration is sufficiently complex to represent a typical motion planning scenario. In practical applications, more complicated robot tasks can be decomposed into a sequence of basic trajectories as such. This will be reflected by the experimental study in Sec.~\ref{sec:exp} which showcases how to apply the developed algorithm to a practical pick-and-place task. 

\subsection{Sampling Domains}

The sampling domain of the goal positions in the test studies can be set the same as the experimental domain defined above to validate the true generalizability of the agents. However, the sampling domain for training should be properly determined since a large training domain tends to cause large variances in the training process, bringing down its stability and convergence speed. Considering this, the sampling domains in this paper are determined as follows.

\subsubsection{Training Domain}

In training studies, each goal position $\x_{\mathrm{g}}$ is uniformly sampled from a polyhedral domain $\mathcal{P}_{\mathrm{g}}=\{P_{\mathrm{g}},\,\Delta P_{\mathrm{g}}\}$ with $P_{\mathrm{g}}=(0.30~0.35~0.08)\,$m being its center coordinate and $P_{\mathrm{g}} \!+\! \Delta P_{\mathrm{g}}$ being its vertexes, where $\Delta P_{\mathrm{g}}\!=\!(\pm 0.05~\pm 0.05~\pm 0.01\,)\,$m. The purpose of uniform sampling is to improve the robustness of the trained agent against the perturbation of the actual goal positions. 
We also randomly sample the obstacle position by $\x_{\mathrm{b}} \!\in\! \mathcal{P}_{\mathrm{b}}(\x_{\mathrm{i}}, \x_{\mathrm{g}})$ to improve the robustness of the agent to obstacle positions, where $\mathcal{P}_{\mathrm{b}}(\x_{\mathrm{i}}, \x_{\mathrm{g}}) \!=\! \{P_{\mathrm{b}}(\x_{\mathrm{i}}, \x_{\mathrm{g}}), \Delta P_{\mathrm{b}}\}$ is a polyhedral sampling region that depends on the initial position $\x_{\mathrm{i}}$ and the sampled goal position $\x_{\mathrm{g}}$, with $\displaystyle P_{\mathrm{b}} = \frac{\x_{\mathrm{i}} \!+\! \x_{\mathrm{g}}}{2}$ being the geometry center and $P_{\mathrm{b}} \!+\! \Delta P_{\mathrm{b}}$ being the vertexes, where $\Delta P_{\mathrm{b}}\!=\!(\pm 0.05~\pm 0.05~\pm 0.02)\,$m. We set the obstacle sampling space $\mathcal{P}_{\mathrm{b}}$ generally in the mid-way between the fixed initial position $\x_{\mathrm{i}}$ and the goal sampling space $\mathcal{P}_{\mathrm{g}}$ to intentionally create challenging configurations for the agent training.

\subsubsection{Test Domain}

In test studies, the goal positions $\x_{\mathrm{g}}$ are sampled from a polyhedral domain $\tilde{\mathcal{P}}_{\mathrm{g}}\!=\!\{\tilde{P}_{\mathrm{g}}, \Delta \tilde{P}_{\mathrm{g}}\}$, where $\tilde{P}_{\mathrm{g}}\!=\!(0.32~0.34~0.09)\,$m and $\Delta \tilde{P}_{\mathrm{g}}\!=\!(\pm 0.3~\pm 0.25~\pm 0.05)\,$m,
i.e., the same as the experimental domain $\mathcal{D}$ defined above. During the sampling of the goal positions, we also eliminate the ones that are too close to the initial positions to guarantee the feasibility of trajectory generation. For each sampled goal position $\x_{\mathrm{g}}$, the obstacle position $\x_{\mathrm{b}}$ is also randomly sampled from a polyhedral region $\tilde{\mathcal{P}}_{\mathrm{b}}=\{\tilde{P}_{\mathrm{b}}, \Delta \tilde{P}_{\mathrm{b}}\}$ of which the center $\displaystyle \tilde{P}_{\mathrm{b}} \!=\! \frac{\x_{\mathrm{i}} \!+\! \tilde{P}_{\mathrm{g}}}{2}$ is the mid-point between the initial position and the goal position, and the vertexes are $\displaystyle \frac{\x_{\mathrm{i}} + \tilde{P}_{\mathrm{g}}}{2} \!+\! \Delta \tilde{P}_{\mathrm{b}}$, where $\Delta \tilde{P}_{\mathrm{b}}\!=\!(\pm 0.05,\,\pm 0.05,\,\pm 0.02)\,$m.

With such settings, the test domain is consistent with the experimental domain. They fully cover and even far exceed the training domain, rendering a large generalization magnitude. This ensures that the test studies can reflect the true generalizability level of the agents.

\subsection{Agents for Ablation and Comparison Studies}

%
The objective of the experiments is to showcase the advantage of IBC in RL-based robot motion planning. To this end, we compare the training and test performance of the proposed IBC-DMP RL method with the following ablated agents.
\begin{itemize}[leftmargin=*]
\item \textit{DDPG-DMP}: based on \textit{IBC-DMP} in Fig.~\ref{fig:over_view}, its demo buffer is removed (IBC ablated) and its replay buffer is initialized with random data generated by uniformly sampled actions, instead of demonstration data (DIR ablated). It serves as a typical DDPG method not incorporating demonstrations;

\item \textit{DEMO-DMP}: based on \textit{IBC-DMP} in Fig.~\ref{fig:over_view}, its demo buffer is removed (IBC ablated) but the replay buffer is still initialized using the demonstration data (DIR reserved); This method serves as a fairly comparable benchmark to validate the advantage of IBC.

\item \textit{EBC-DMP}: based on \textit{IBC-DMP} in Fig.~\ref{fig:over_view}, its BC Loss in Fig.~\ref{fig:over_view} is computed using an EBC method as in Eq.~\eqref{eq:bc_loss_old}, instead of IBC; This method is intended to showcase the advantage of IBC over EBC for an RL agent.
\end{itemize}

For a fair comparison, the agents are configured with the same parameters, as shown in Tab.~\ref{tab:hyper-param}. They also use neural networks with the same structures, namely a three-layer structure of which the neuron numbers are $64$, $128$, and $64$, respectively. The activation functions of the networks are ReLU functions. The weights and biases of the networks are randomly initialized. Besides, the training parameters for the IBC-DMP agent and the EBC-DMP agent are shown in Tab.~\ref{tab:hyper-param-II}. Note that these parameters have been fine-tuned through empirical studies which are not elaborated in detail in this paper, for brevity.

\linespread{1.5}
\begin{table}[htbp]
\centering
\caption{Training parameters of all agent}
\begin{tabular}{c|c|c|c|c|c}
	\thickhline
	\rowcolor{dark_purple} {\color{white} \textbf{Par.}} & {\color{white} \textbf{Name}} & {\color{white} \textbf{Val.}} & {\color{white} \textbf{Par.}} & {\color{white} \textbf{Name}} & {\color{white} \textbf{Val.}} \\
        \hline
        $\gamma$ & reward discount & $0.99$ & $\sigma^2$ & action variance & $0.01$ \\
        \rowcolor{shadow_purple} $\lambda$ &
        target factor & $0.995$ & $\alpha$ & learning rate & $10^{-4}$ \\
        $T$ & steps per episode & $250$ & $\Delta t$ & time per step & $0.02\,$s \\
        \rowcolor{shadow_purple} $N$ & number episode & $500$ & $N_{\mathrm{R}}$ & replay size & $10^6$ \\
	\thickhline
\end{tabular}
\label{tab:hyper-param}
\end{table}
\linespread{1}

\linespread{1.5}
\begin{table}[htbp]
\centering
\caption{Training parameters of IBC/EBC-DMP}
\begin{tabular}{c|c|c|c|c|c}
	\thickhline
	\rowcolor{dark_purple} {\color{white} \textbf{Par.}} & {\color{white} \textbf{Name}} & {\color{white} \textbf{Val.}} & {\color{white} \textbf{Par.}} & {\color{white} \textbf{Name}} & {\color{white} \textbf{Val.}} \\
        \hline
        $\lambda_{\mathrm{BC}}$ & BC ratio & $2$ & $N_{\mathrm{D}}$ & demo size & $5 \times 10^4$ \\
        \rowcolor{shadow_purple} $N_{\mathrm{D}}^Q$ & batch size $\mathcal{B}_{\mathrm{D}}^{Q}$ & $450$ & $N_{\mathrm{R}}^Q$ & batch size $\mathcal{B}_{\mathrm{R}}^{Q}$ & $50$ \\
        $N_{\mathrm{D}}^{\pi}$ & batch size $\mathcal{B}_{\mathrm{D}}^{\pi}$ & $100$ &
        $N_{\mathrm{R}}^{\pi}$ & batch size $\mathcal{B}_{\mathrm{R}}^{\pi}$ & $500$ \\
	\thickhline
\end{tabular}
\label{tab:hyper-param-II}
\end{table}
\linespread{1}

\subsection{Training Performance}\label{sec:sim_training}

This subsection evaluates the training performance of the agents using random goal and obstacle positions sampled from the training domain $\mathcal{P}_{\mathrm{g}}$ introduced in Sec.~\ref{sec:domains}.
For each agent, we train $10$ policies with the same initial condition but with different random seeds. The evaluation of the training performance of the agents will be conducted on all 10 policies to balance the effects of randomness.

\subsubsection{Training Scores}\label{sec:Lij}
\begin{figure}[htbp]
    \centering
    \subfloat[\footnotesize{IBC-DMP}]{\includegraphics[width=0.23\textwidth]{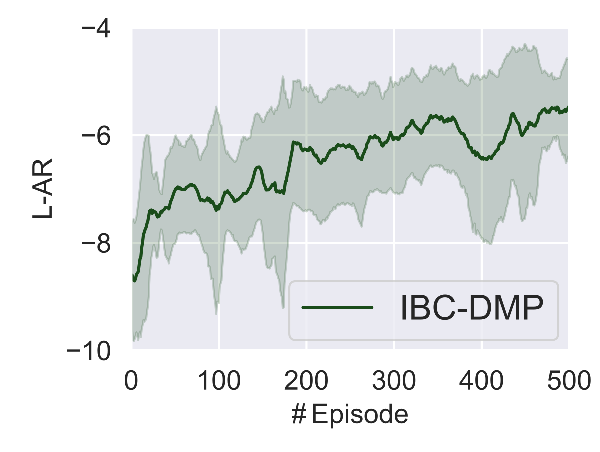}
        \label{fig:ibc_training}}
    \hfill
    \subfloat[\footnotesize{DEMO-DMP}]{\includegraphics[width=0.23\textwidth]{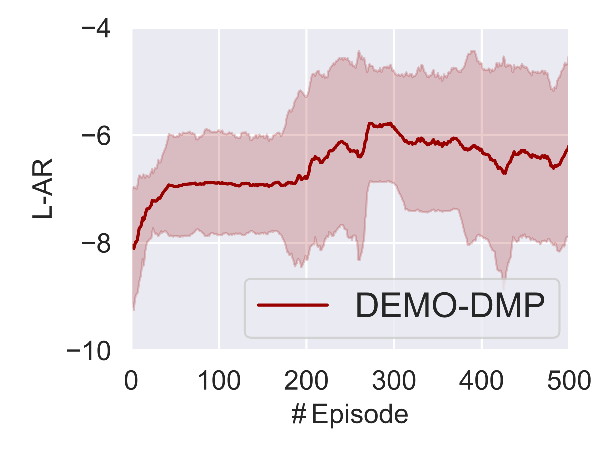}
        \label{fig:bc_a_training}}
    \hfill
    \subfloat[\footnotesize{DDPG-DMP}]{\includegraphics[width=0.23\textwidth]{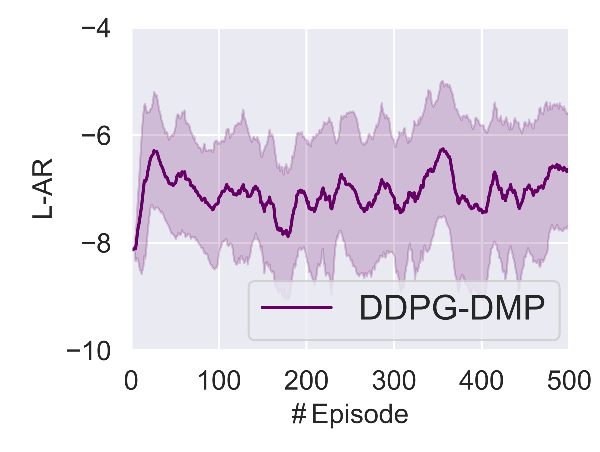}
        \label{fig:demo_a_training}}
        \hfill
    \subfloat[\footnotesize{EBC-DMP}]{\includegraphics[width=0.23\textwidth]{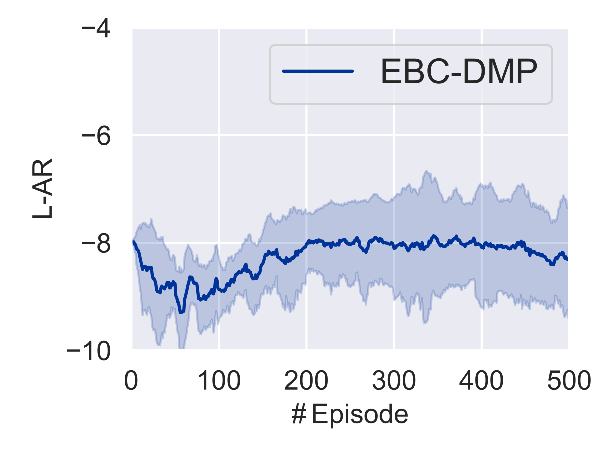}
        \label{fig:ebc_training}}
    \caption{The L-AR lines of the agents in training.}
    \label{fig:bc_training}
\end{figure}

For each episode $i=1,2,\cdots, N$, we use the logarithm of the accumulated reward (L-AR) score, defined as $L_i\!=\!-\mathrm{log}(1\!-\!\sum_{t=0}^Tr_t^{i})$, to value the training performance of an agent, where $r_t^{i}$ is the instant reward at step $t=0,1,\cdots, T$ of episode $i$. Then, the performance of an RL agent can be evaluated by whether the L-AR score reaches a high value. Fig.~\ref{fig:bc_training} shows the change in L-AR scores as the episode increases. In each subfigure, the solid line and the shadow region denote the mean values and the standard deviations of the L-AR score over all random runs, respectively. We smooth out the lines in all subfigures with $10$ episodes for clear presentation.

The training performance of the ultimate policy can be quantified using the ultimate training score $L_{N}$. Tab.~\ref{tab:qua_metric} shows the averaged value $\mu(L_{N})$ and the standard deviation $\sigma(L_{N})$ of the ultimate L-AR scores of the four agents. 

\linespread{1.5}
\begin{table}[htbp]
\centering
\caption{The training scores}
\begin{tabular}{c|c|c|c|c}
\thickhline
\rowcolor{dark_red}
{\color{white}\textbf{Scores}} & {\color{white}\textbf{IBC-DMP}} & {\color{white}\textbf{DEMO-DMP}} & {\color{white}\textbf{DDPG-DMP}} & {\color{white}\textbf{EBC-DMP}}\\
\hline
$\mu(L_{N})$ & \textbf{-5.5415} & \textbf{-6.2713} & \textbf{-6.6617} & \textbf{-8.2584} \\
\rowcolor{shadow_red} $\sigma(L_{N})$ & \textbf{1.0542} & \textbf{1.5261} & \textbf{1.0423} & \textbf{0.7677} \\
\thickhline
\end{tabular}
\label{tab:qua_metric}
\end{table}
\linespread{1}

\subsubsection{Training Performance}

In Fig.~\ref{fig:bc_training}, the L-AR scores of all agents start from an approximate level $-8$ which can be seen as a perform baseline. From Fig.~\ref{fig:ibc_training}, we can see that the L-AR score of the IBC-DMP agent increases steadily, ultimately reaching a high value at the $500$-th episode. This validates the efficacy of IBC-DMP in terms of fast training. Moreover, IBC-DMP ultimately achieves the highest average L-AR score compared to other agents, as reflected by both Fig.~\ref{fig:ibc_training} and Tab.~\ref{tab:qua_metric}, showing its superior learning performance. 

In the meantime, DEMO-DMP and DDPG-DMP behave worse than IBC-DMP due to the ablation of IBC. The DDPG-DMP agent does not show a significant increase in L-AR scores during the learning process (Fig.~\ref{fig:demo_a_training}). This reflects the drawback of the DDPG algorithm in terms of slow learning due to the bootstrapping-based value approximation. Compared to DDPG-DMP, the DEMO-DMP agent shows an obvious increase in L-AR scores (Fig.~\ref{fig:demo_a_training}) and reaches a higher score ultimately (Tab.~\ref{tab:qua_metric}). This implies that the DIR technology can effectively promote the learning of an RL agent. Nevertheless, Tab.~\ref{tab:qua_metric} also shows that DEMO-DMP leads to a larger standard deviation than DDPG-DMP due to the diverse distribution between the demonstration data and the experience data. Therefore, we can conclude that the DIR technology can improve the learning speed of an RL agent while bringing down the stability. IBC serves as an effective cure that further improves learning performance while ensuring learning stability. In this sense, IBC and DIR should be used together for decent learning performance.

The EBC-DMP agent shows the worst training performance among all agents, although it has the best stability reflected by the smallest standard deviation of L-AR scores. Its ultimate average L-AR is even lower than the performance baseline $-8$.  This clearly reflects the overfitting of an EBC-based agent to human demonstrations. The comparison showcases that EBC may be beneficial for an agent to imitate the demonstration, but not suitable for RL training with predefined rewards.

\subsection{Visualized Test with Fixed Goals}

\begin{figure}[htbp]
    \centering
    \subfloat[\footnotesize{IBC-DMP}]{\includegraphics[width=0.22\textwidth]{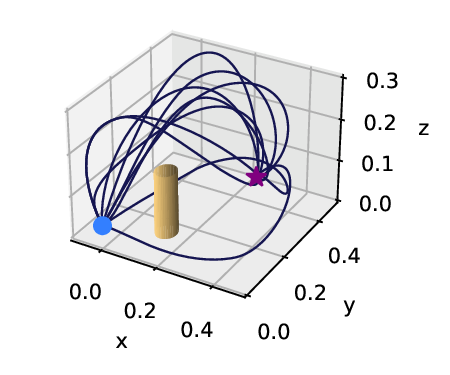}
        \label{fig:bc_test_0}}
    \hfill
    \subfloat[\footnotesize{DEMO-DMP}]{\includegraphics[width=0.22\textwidth]{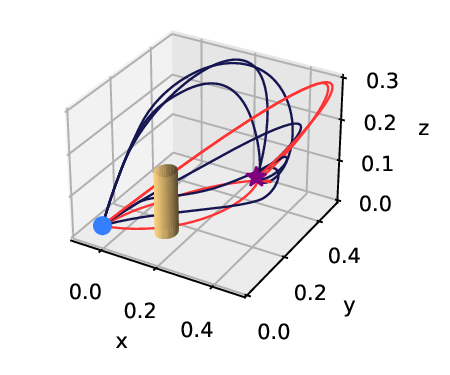}
        \label{fig:bc_test_1}}
    \hfill
    \subfloat[\footnotesize{DDPG-DMP}]{\includegraphics[width=0.22\textwidth]{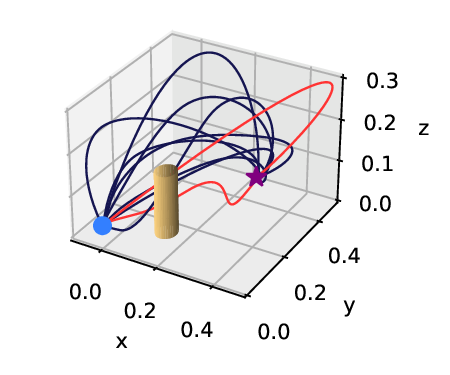}
        \label{fig:bc_test_1_5}}
    \hfill
    \subfloat[\footnotesize{EBC-DMP}]{\includegraphics[width=0.22\textwidth]{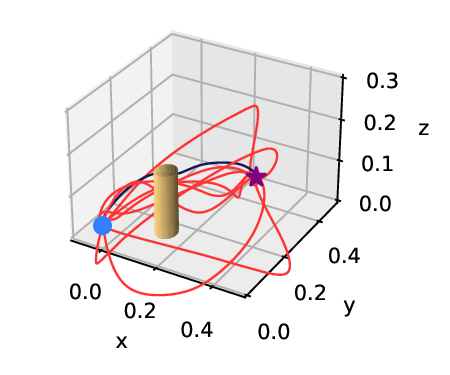}
        \label{fig:bc_test_2}}
    \caption{The test trajectories of the agents for fixed goal and obstacle positions, where the blue dot is the initial position, the purple star denotes the goal position, and the yellow cylinder represents the obstacle. The trajectories that collide with the obstacle or with the ground are marked as red, while those that do not are in blue.}
    \label{fig:bc_test}
\end{figure}

This subsection uses a visualized manner to showcase the intuitive performance of the trained agents with a fixed goal position $\x_{\mathrm{g}}\!=\!(0.32~0.34~0.09)\,$m and an obstacle position $\x_{\mathrm{b}}\!=\!(0.13~0.14~0.16)\,$m. The resulting trajectories of the trained agents for given goal and obstacle positions are illustrated in Fig.~\ref{fig:bc_test}. Each subfigure shows the trajectories of the corresponding agent generated from $10$ trained policies. The red trajectories are those colliding with the obstacle or with the ground. Here, we refer to a trajectory as \textit{with collisions} if it intersects with the obstacle or with the ground, i.e., if it has at least $1$ discrete-time samples that are within the cylinder obstacle domain or under the ground. It is noticed that the IBC-DMP agent achieves zero collisions while each other agent has at least two trajectories with collisions. This indicates the superior performance of the IBC-DMP agent in terms of collision avoidance in motion planning tasks.

Note that Fig.~\ref{fig:bc_test} only provides an intuitive result to showcase the essential performance of the trained agents for simple motion planning tasks, not yet sufficient to validate their real generalizability. The test studies with sufficient sampling from the test domains to be interpreted in the following subsection will cover the generalizability validation of the agents.

\subsection{Test Performance for Random Goals}\label{sec:fix_goal}

This subsection evaluates the test performance of the trained agents with random goal and obstacle positions sampled from the test domain $\tilde{\mathcal{P}}_{\mathrm{g}}$ introduced in Sec.~\ref{sec:domains}. For each trained policy, we perform $100$ test runs with different random goal positions $\x_{\mathrm{g}}$ and random obstacle positions $\x_{\mathrm{b}}$, leading to $N_{\mathrm{t}}=1000$ test runs per agent. The purpose of this study is to provide comprehensive scores to quantify the generalizability of these agents.

\subsubsection{Test Scores}

We also use the L-AR score defined in Sec.~\ref{sec:sim_training} to quantify the test performance of the agents, i.e., $L\!=\!-\mathrm{log}(1\!-\!\sum_{t=0}^Tr_t)$, where $r_t$ is the instant reward at step $t=0,1,\cdots, T$ of a test run. The mean value and the standard deviation of the L-AR scores are listed in Tab.~\ref{tab:test_score}.

\linespread{1.5}
\begin{table}[htbp]
\centering
\caption{The test scores}
\begin{tabular}{c|c|c|c|c}
	\thickhline
\rowcolor{dark_blue}	{\color{white} \textbf{Scores}} & {\color{white} \textbf{IBC-DMP}} & {\color{white} \textbf{DEMO-DMP}} & {\color{white} \textbf{DDPG-DMP}} & {\color{white} \textbf{EBC-DMP}} \\
        \hline
$\mu(L)$ & \textbf{-6.1439} & \textbf{-6.2848} & \textbf{-6.2612} & \textbf{-8.2068} \\
\rowcolor{shadow_blue} $\sigma(L)$ & \textbf{0.8654} & \textbf{1.2455} & \textbf{0.6189} & \textbf{1.1432}  \\
	\thickhline
\end{tabular}
\label{tab:test_score}
\end{table}
\linespread{1}

We also use the collision rate defined as $R_{\mathrm{c}} \!=\! N_{\mathrm{c}}/N_{\mathrm{t}}$ to evaluate the collision avoidance performance of an agent in the test, where $N_{\mathrm{c}}$ is the number of trajectories with collisions. Tab.~\ref{tab:collision_rates} lists the mean value and the standard deviation of the collision rates over all random runs, for all four agents. 

\linespread{1.5}
\begin{table}[htbp]
\centering
\caption{The collision rates}
\begin{tabular}{c|c|c|c|c}
	\thickhline
\rowcolor{dark_yellow}	{\color{white} \textbf{Scores}} & {\color{white} \textbf{IBC-DMP}} & {\color{white} \textbf{DEMO-DMP}} & {\color{white} \textbf{DDPG-DMP}} & {\color{white} \textbf{EBC-DMP}} \\
        \hline
$\mu(R_{\mathrm{c}})$ & \textbf{13.7\%} & \textbf{17.6\%} & \textbf{30.3\%} & \textbf{89.5\%} \\
\rowcolor{mid_yellow} $\sigma(R_{\mathrm{c}})$ & \textbf{0.2641} & \textbf{0.3116} & \textbf{0.3326} & \textbf{0.1595}  \\
	\thickhline
\end{tabular}
\label{tab:collision_rates}
\end{table}
\linespread{1}

\subsubsection{Test Performance}

Tabs.~\ref{tab:test_score} and~\ref{tab:collision_rates} have clearly shown that IBC-DMP presents the highest average score and the lowest collision rate among all agents, indicating the superior generalizability of IBC-DMP in terms of L-AR scores and collision avoidance. 

Compared to DDPG-DMP, DEMO-DMP presents inferior test performance (lower $\mu(L)$ and larger $\sigma(L)$) but superior collision avoidance performance (lower $\mu(R_{\mathrm{c}})$ and smaller $\sigma(R_{\mathrm{c}})$). Its inferior test performance is due to the overfitting to the diverse training data brought up by the mixture of the demonstration data and the experience data. Nevertheless, the diversity of the training data can effectively improve collision avoidance performance. This is because the demonstration data have good performance of collision avoidance, even though it may not align with the desired reward. Based on this, IBC can improve both the test performance and the collision avoidance performance of DEMO-DMP.

Similar to training studies, EBC-DMP shows the lowest average scores and collision rates, although with decent standard deviations. This implies the drawback of EBC in terms of overfitting human actions.

\section{Hardware Experiments}\label{sec:exp}

\begin{figure}[htbp]
    \centering
    \subfloat[]{
    \begin{tikzpicture}
    \node[inner sep=2pt] (camera2) at (3.5cm, 1.9cm])
    {\includegraphics[height=4.7cm]{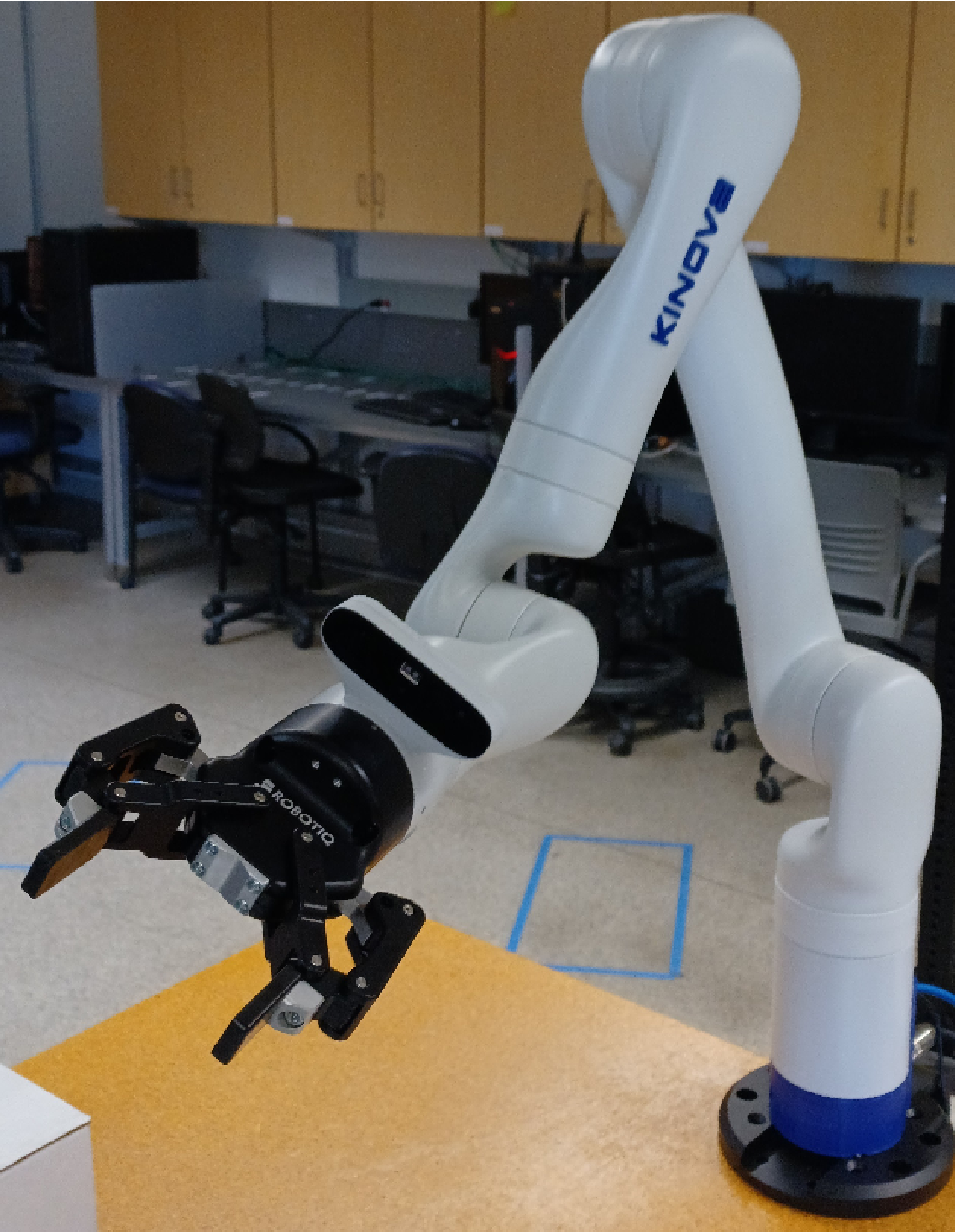}};
    \end{tikzpicture}
        \label{fig:kinova_robot}}
    \hspace{-0.3cm}
    \subfloat[]{
    \begin{tikzpicture}
    \node[inner sep=2pt] (camera2) at (3.5cm, 1.9cm])
    {\includegraphics[height=4.7cm]{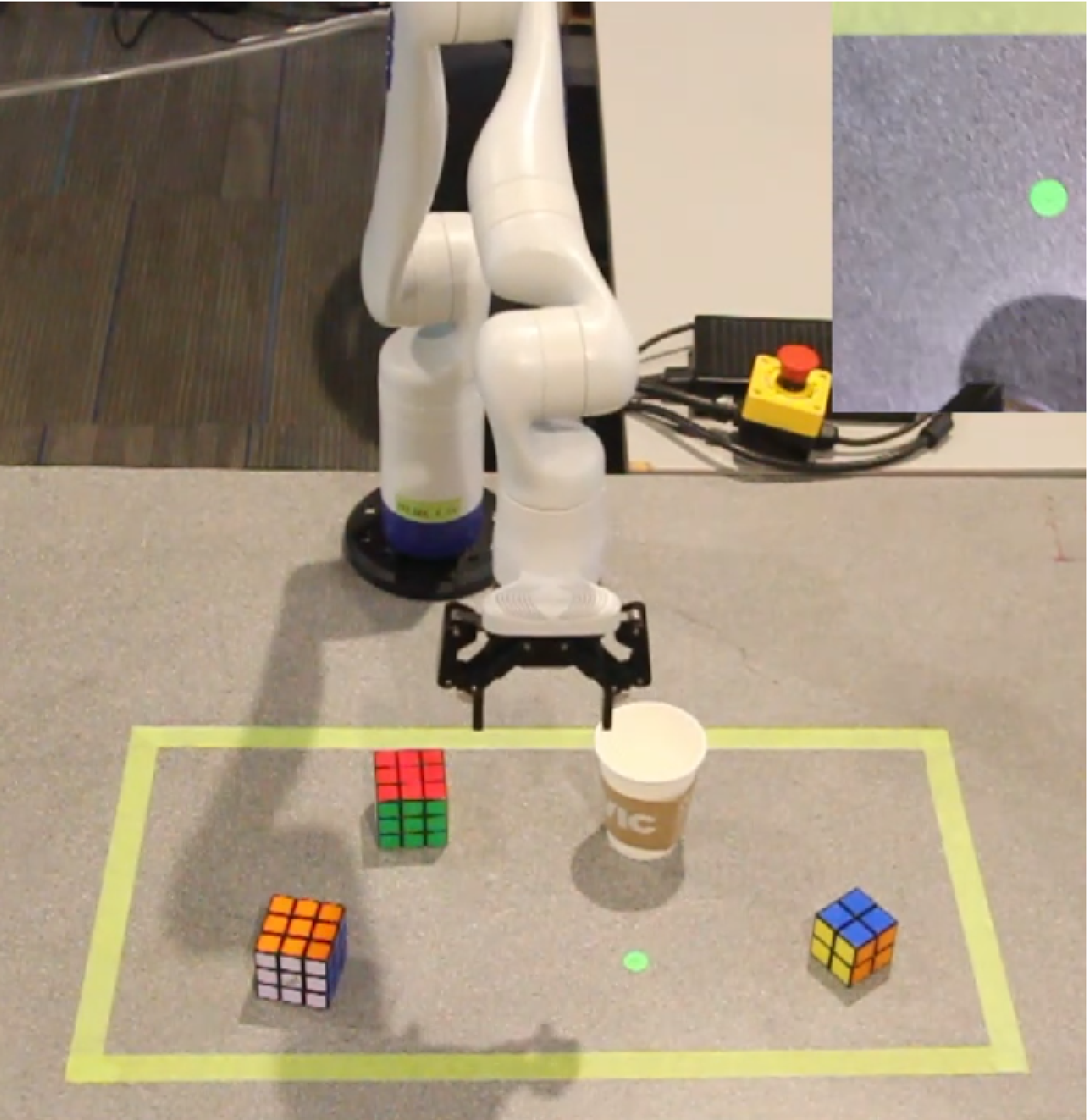}};
    \node[circle,inner sep=0pt,minimum size=1mm,draw, fill=black] (origin) at ([xshift=1.9cm, yshift=2.5cm] camera2.south west) {};
    \draw[->, >=Stealth,color=red, very thick] (origin.center) -- node[pos=1, anchor=south east]{$\pmb{x}$} ([xshift=-0.05cm, yshift=-0.95cm] origin.center);
    \draw[->, >=Stealth,color=blue, very thick] (origin.center) --node[pos=1, anchor=north]{$\pmb{y}$} ([xshift=1cm] origin.center);
    \draw[->, >=Stealth,color=green, very thick] (origin.center) -- node[pos=1, anchor=north west]{$\pmb{z}$} ([yshift=1cm] origin.center);
    \end{tikzpicture}
        \label{fig:robot_2}}
    \caption{Kinova$^\circledR$ Gen3 robot (a) and the assembly scenario (b).}
    \label{fig:robots}
\end{figure}

In this section, we use an experimental case study to evaluate the performance of the proposed IBC-DMP RL method. A Rubik's cube-stacking case is used to demonstrate how to use an IBC-DMP agent to accomplish a general pick-and-place-based assembly task. The robot used in this study is a 6-DoF Kinova\textregistered~Gen$\,$3 manipulator as shown in Fig.~\ref{fig:kinova_robot}, equipped with a two-finger Robotiq$^\circledR$ 2F-85 gripper and an Omnivision OV5640 on-gripper camera. The robot is connected to a desktop workstation with an AMD\textregistered~Ryzen9 3950X CPU, an Intel\textregistered~RTX4090 GPU, and a Ubuntu 18.04 operating system. The robot is controlled by the Kinova$^\circledR$ Kortex Python API in https://github.com/Kinovarobotics/kortex. An RGB camera in front of the robot provides a third-person perspective, as shown in Fig.~\ref{fig:robot_2}. Footage of the hardware experiment is available at https://youtu.be/o9tD9N-rEOc?si=exzk3wDX4pJabSkM.

\subsection{Experimental Configuration}

As shown in Fig.~\ref{fig:robot_2}, the workspace of the cube-stacking scenario is marked as a rectangular region on the table in front of the Gen3 robot arm. We set the origin of the task coordinate at the robot base. The axes of the coordinate are shown as colored arrows in Fig.~\ref{fig:robot_2}. The coordinates of the vertexes of the workspace are (0.25,\,0.45,\,0)\,m, (0.25,\,-0.2,\,0)\,m, (0.6,\,0.45,\,0)\,m, and (0.6,\,-0.2,\,0)\,m, respectively. The size of the workspace is determined to cover the largest view of the on-gripper camera within the reach of the robot gripper. Three Rubik's cubes with different sizes (two cubes with a 57 mm edge and one cube with a 45 mm edge) and colors are randomly placed within the rectangular region manually. A paper cup with a radius of 90 mm and a height of 110 mm is manually placed at a random position as an obstacle. As shown in Fig.~\ref{fig:robot_2}, the robot is commanded to start from the HOME position at (0.356, 0.106, 0.277)\,m, pick up the cubes, and place them at the GOAL position (marked as a green light point) one by one in the order of red, orange, and blue, until they are piled up in a column. The piling-up process can be recognized as a simple assembly task, during which the robot gripper should not collide with the paper cup. The vision of the cubes and the cup obstacle is captured using the on-gripper camera and their ground-truth positions are calculated using the pre-built object detection libraries in YOLOv8~\cite{Jocher_YOLO_by_Ultralytics_2023} and OpenCV~\cite{opencv_library}. 

For each experimental trial, we represent the initial positions of the three cubes with red, orange, and blue top surfaces as $P_{\mathrm{red}}$, $P_{\mathrm{org}}$, and $P_{\mathrm{blu}}$, respectively. We also set three via-points $\tilde{P}_{\mathrm{red}}$, $\tilde{P}_{\mathrm{org}}$, and $\tilde{P}_{\mathrm{blu}}$ 5\,cm right above them to ease the grasping of the cubes. In this sense, six trajectories need to be generated for the cube-stacking task, i.e., HOME\,--\,$\tilde{P}_{\mathrm{red}}$\,--\,$G_{\mathrm{red}}$\,--\,$\tilde{P}_{\mathrm{org}}$\,--\,$G_{\mathrm{org}}$\,--\,$\tilde{P}_{\mathrm{blu}}$\,--\,$G_{\mathrm{blu}}$, where $G_{\mathrm{red}}$, $G_{\mathrm{org}}$, and $G_{\mathrm{blu}}$ are the stacking positions of the cubes in the ultimate stacking column. The procedure of the stacking task is described in Algorithm~\ref{ag:task}, where \textit{DMP} is a function used to generate the desired trajectories using the multi-DoF DMP model \eqref{eq:dmp_multi_dof} with given initial, goal, and obstacle positions $\x_{\mathrm{i}}$, $\x_{\mathrm{g}}$, and $\x_{\mathrm{b}}$, respectively, and the trained policy $\mathbf{f}$. In this experiment, we do not involve additional task-level complexities since the evaluation is in the trajectory level. Also, the multi-DoF DMP model does not consider the orientations of the gripper for brevity. Instead, the orientations of the gripper are commanded with the following intuitive principles.
\begin{itemize}
\item The gripper always points down to the table, meaning that its $x$- and $y$-orientations are fixed.
\item The $z$-orientations of the gripper at the via points are manually determined to allow successful grasping.
\item The $z$-orientations of the gripper at any non-via points are interpolated linearly.
\end{itemize}
Such a design ensures the successful execution of the cube-stacking task without causing singular configurations. All trajectories are designed in the Cartesian space and are mapped to the joint space of the robot using the pre-built inverse kinematics (IK) library of the Kinova$^\circledR$ Gen$\,$3 robot.

\begin{algorithm}[htbp]
\caption{Cube stacking task procedure}\label{ag:task}
\begin{algorithmic}[1]
\STATE Initialize robot gripper at HOME position
\STATE Assign the paper cup position to $\x_{\mathrm{b}}$
\FOR{$j$ \textbf{in} $\{\mathrm{red}, \mathrm{org}, \mathrm{blu}\}$}
    \STATE $\x_{\mathrm{i}} \leftarrow \x_t$, $\x_{\mathrm{g}} \leftarrow \tilde{P}_{j}$
    \STATE Generate trajectory $(\x_t, \dot{\x}_t)$\,=\,\textit{DMP}($\x_{\mathrm{i}}$, $\x_{\mathrm{g}}$, $\x_{\mathrm{b}}$, $\mathbf{f}$) \label{code:gen_trj}
    \label{code:start}
    \WHILE{$\|\x_t\!-\!\x_{\mathrm{g}}\|\!>\!\varepsilon_N$}
    \STATE Follow trajectory $(\x_t, \dot{\x}_t)$
    \ENDWHILE \label{code:end}
    \STATE Grasp cube at $P_j$ \label{code:grasp}
    \STATE Determine stacking position $\x_{\mathrm{i}} \leftarrow \x_t$, $\x_{\mathrm{g}} \leftarrow G_j$
    \STATE Repeat line \ref{code:start} to line \ref{code:end}
    \STATE Release cube
\ENDFOR

\end{algorithmic}
\end{algorithm}

\subsection{Experiment Results}\label{sec:experi}

\begin{figure*}[htbp]
    \centering
    \subfloat[]{\includegraphics[width=0.22\textwidth]{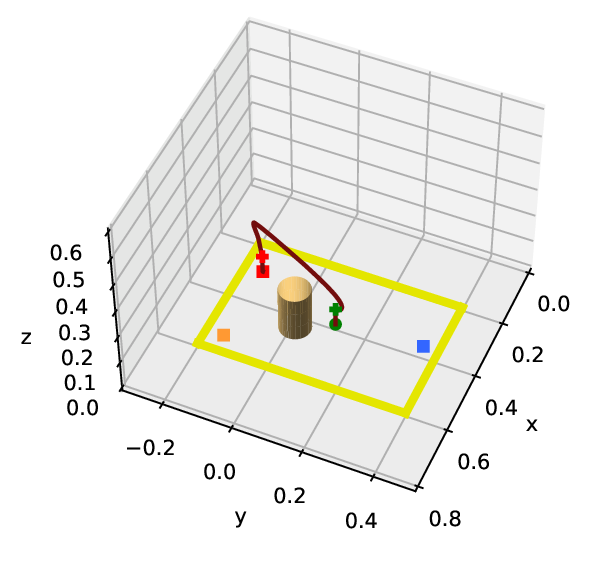}
        \label{fig:exp_3d_1}}
    \hfill
    \subfloat[]{\includegraphics[width=0.22\textwidth]{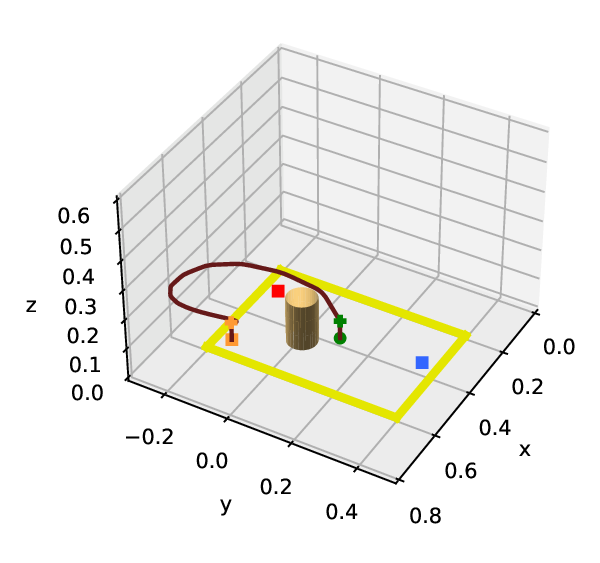}
        \label{fig:exp_3d_2}}
    \hfill
    \subfloat[]{\includegraphics[width=0.22\textwidth]{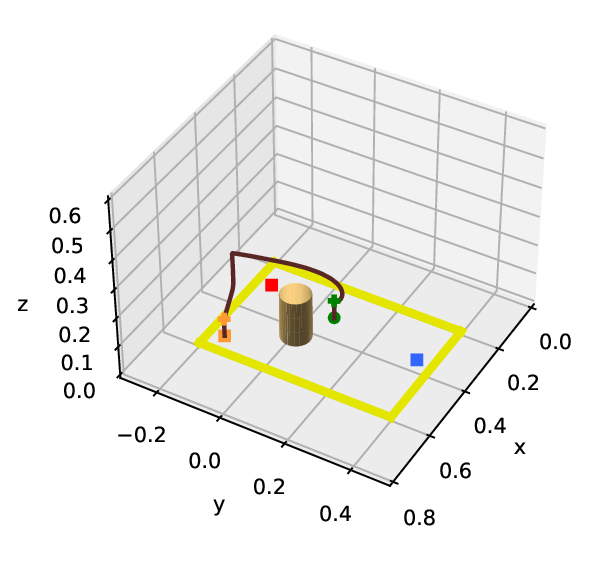}
        \label{fig:exp_3d_3}}
    \hfill
    \subfloat[]{\includegraphics[width=0.22\textwidth]{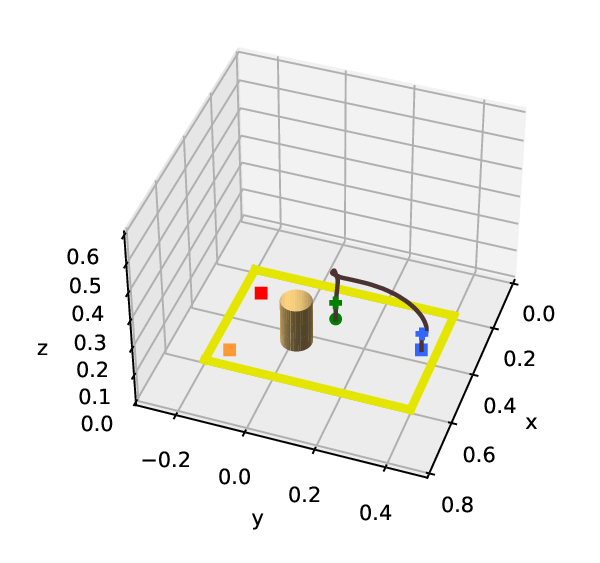}
        \label{fig:exp_3d_4}}
    \hfill
    \subfloat[]{\includegraphics[width=0.22\textwidth]{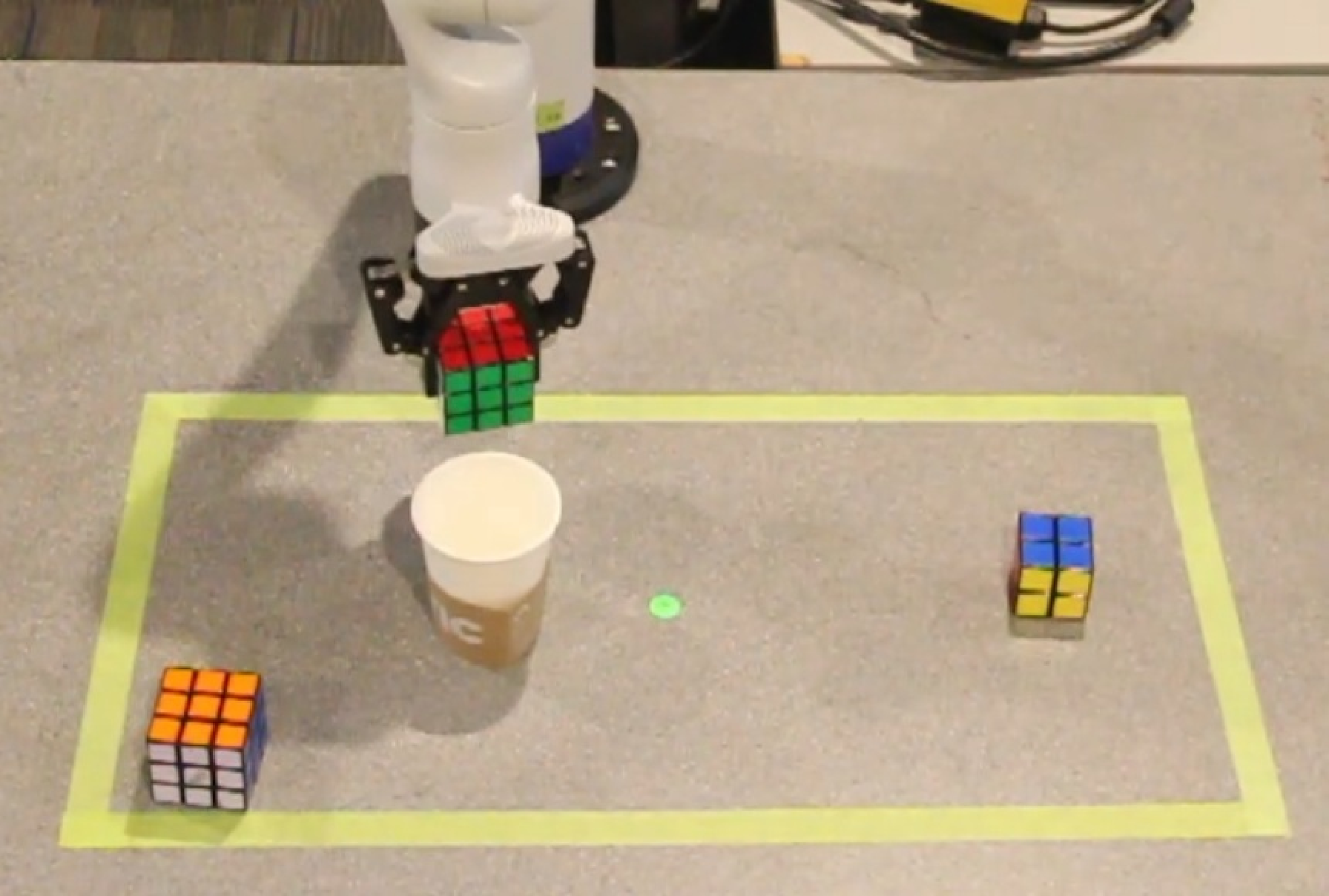}
        \label{fig:video_1}}
        \hfill
    \subfloat[]{\includegraphics[width=0.22\textwidth]{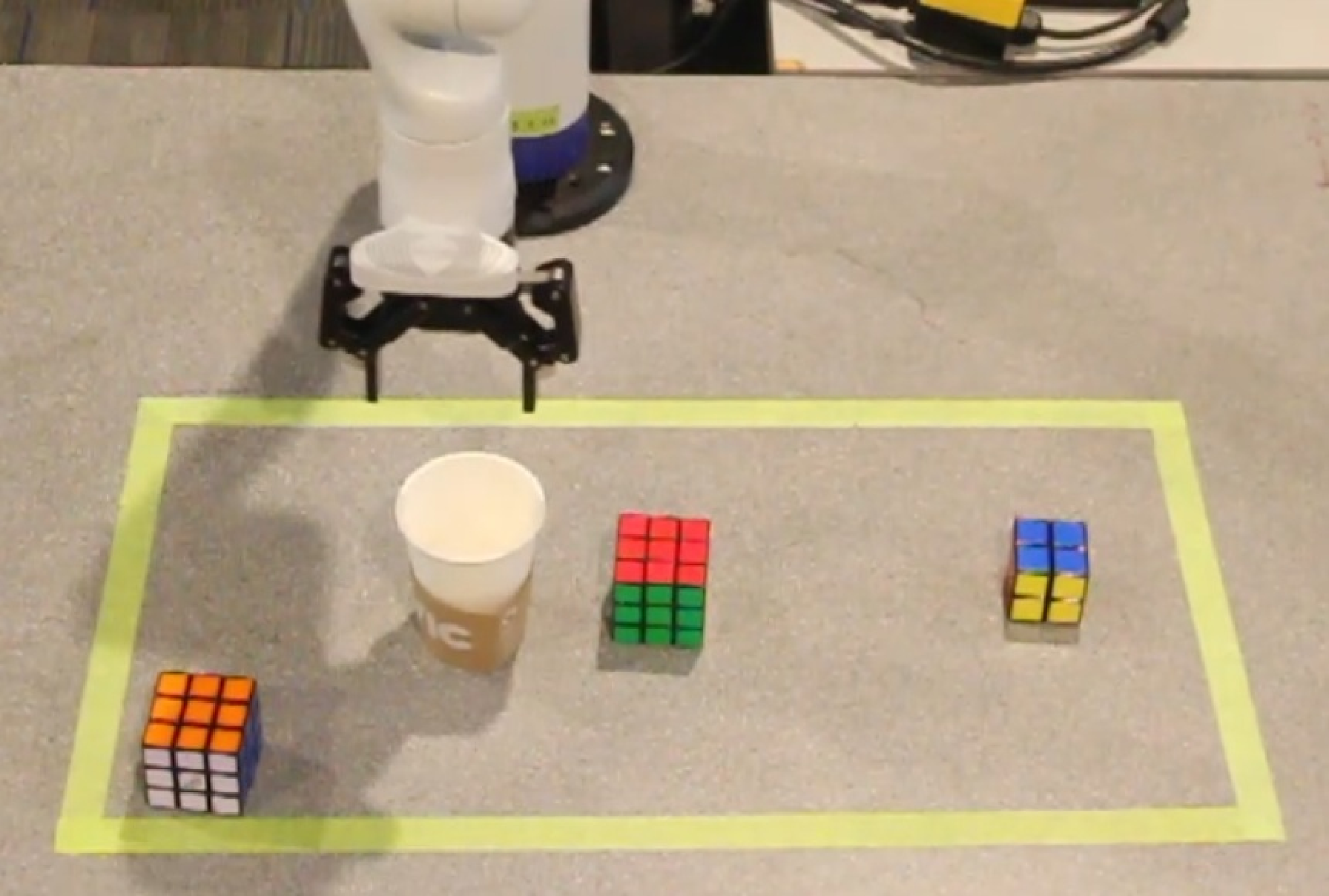}
        \label{fig:video_2}}
    \hfill
    \subfloat[]{\includegraphics[width=0.22\textwidth]{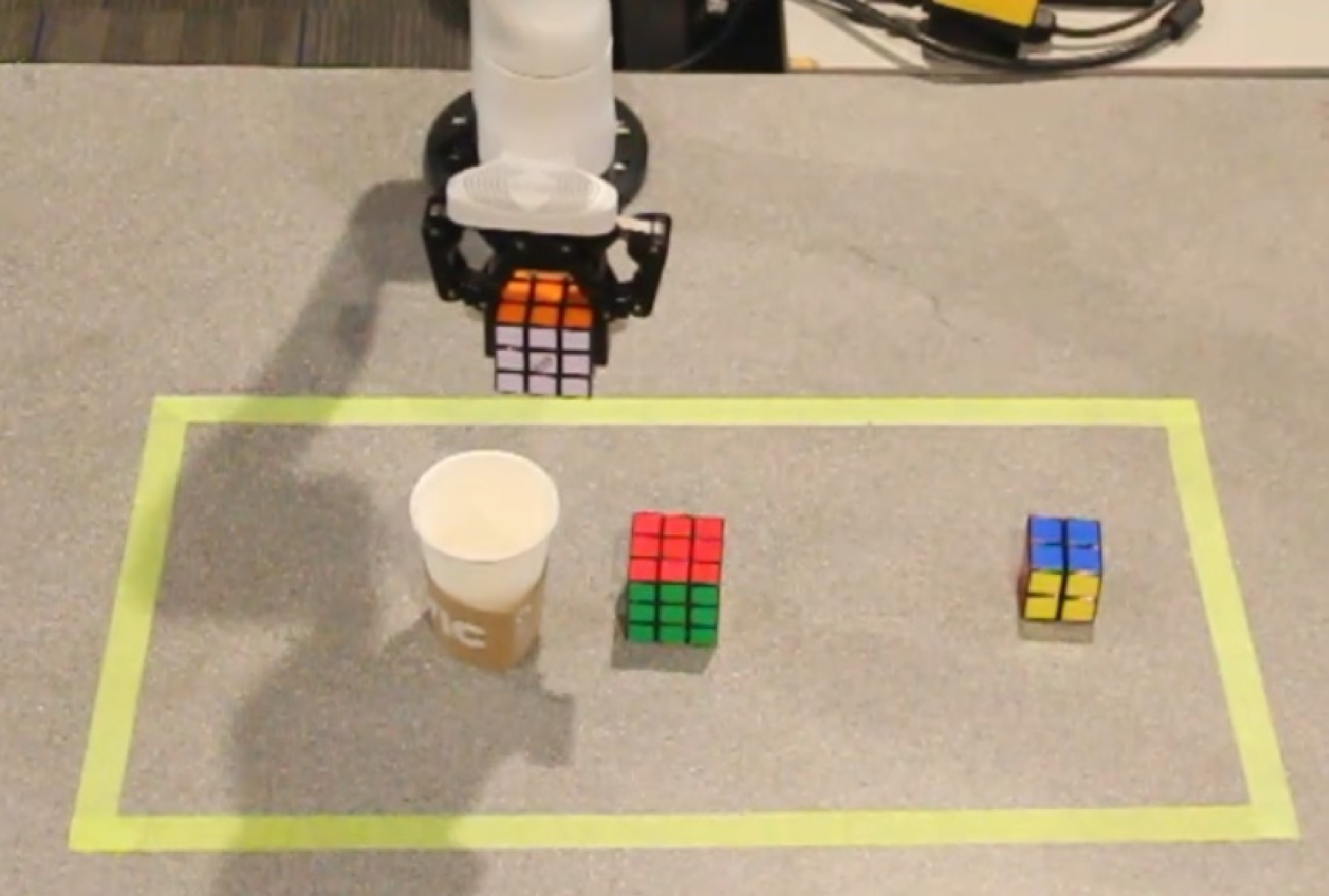}
        \label{fig:video_3}}
    \hfill
    \subfloat[]{\includegraphics[width=0.22\textwidth]{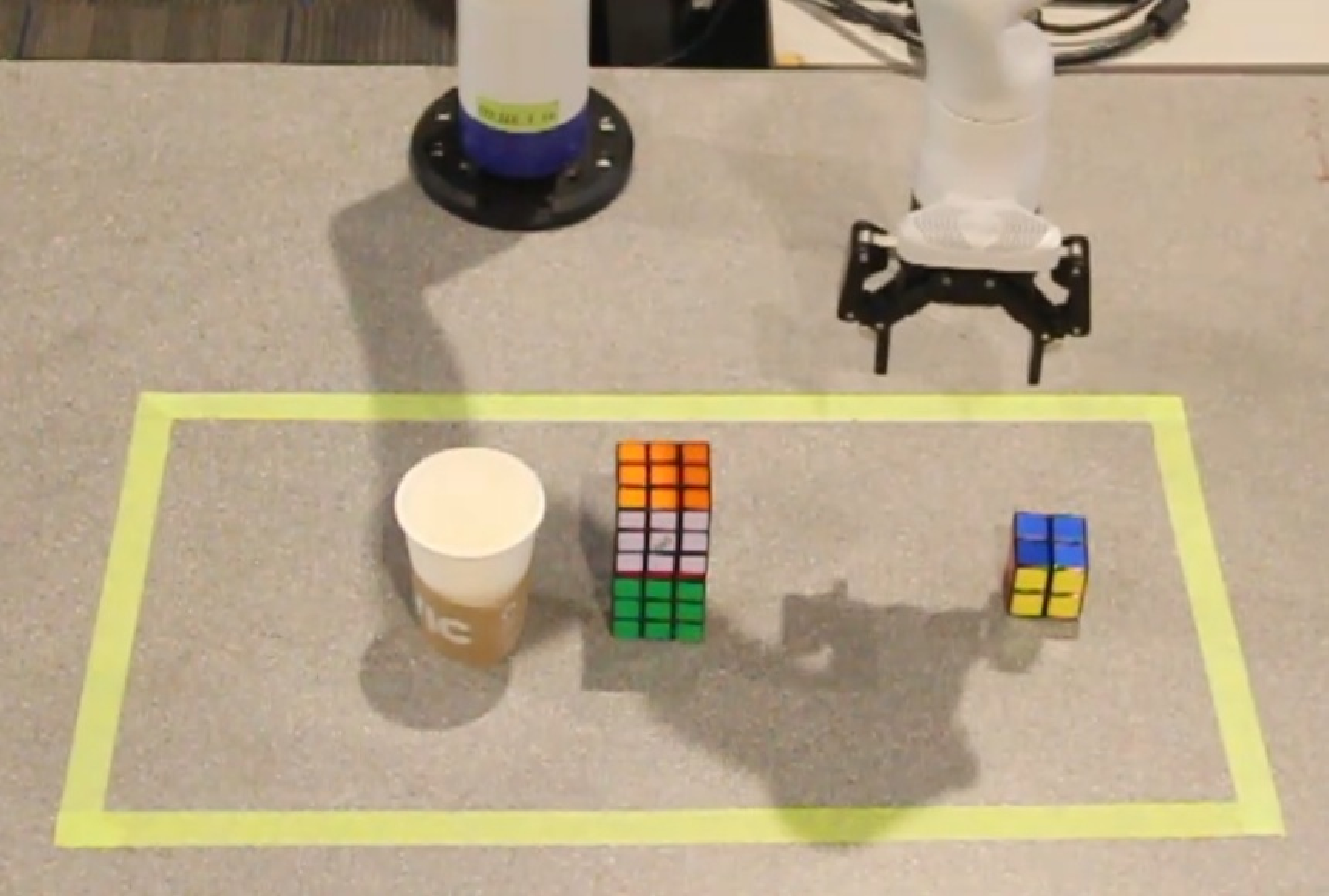}
        \label{fig:video_4}}
    \caption{The 3D visualization and vision of the hardware experiment.}
    \label{fig:exp_results}
\end{figure*}

For an IBC-DMP RL policy trained in Sec.~\ref{sec:va_sim}, we performed 22 experimental trials to incorporate the influence of randomness. For each trial, the initial positions of the cubes and the cup are randomly placed by hand. Also, we intentionally create challenging situations where the robot gripper has a large chance to go around the paper cup. One example trial is shown in Fig.~\ref{fig:exp_results}, where the paper cup is the mid-way between the red and the orange cubes, and the goal position. In this situation, the gripper must go around the cup to avoid collision with it, leading to nontrivial trajectories. Fig.~\ref{fig:exp_results} shows that the IBC-DMP RL policy can successfully generate collision-free trajectories for the robot gripper, which indicates the efficacy of the IBC-DMP RL motion planning framework.

\begin{figure}[htbp]
    \centering
    \includegraphics[width=0.4\textwidth]{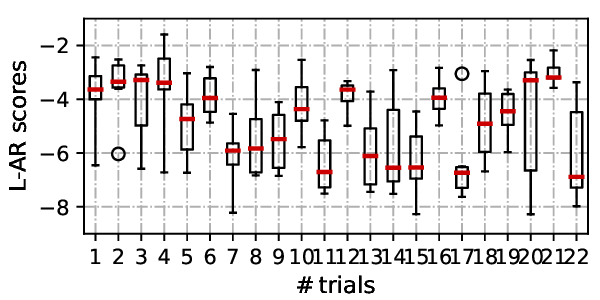}
    \caption{The L-AR scores among different trials.}
    \label{fig:box_plot}
\end{figure}

Similar to Sec.~\ref{sec:fix_goal}, we also use L-AR scores to quantitatively evaluate the performance of the IBC-DMP RL policy in the experimental study. We use the internal position sensor of the Gen 3 robot to record the actual executed paths of the gripper during the cube-stacking task. Fig.~\ref{fig:box_plot} is the box plot of the L-AR scores of these paths grouped according to trials. Each box denotes the distribution of the L-AR scores of the 6 trajectories of an experimental trial. The red bar in each box represents the median value of the L-AR scores, representing the overall performance of the policy in the corresponding trial. It is noticed that all red bars are higher than $-7$, much higher than the performance baseline $-8$ given in Sec.~\ref{sec:sim_training}, showing its efficacy with high scores. Also, the scores of 16 out of 22 trials are higher than the performance standard $-6$, which indicates the decent overall performance of the selected policy concerning L-AR scores. 

\begin{figure}[htbp]
    \centering
    \includegraphics[width=0.4\textwidth]{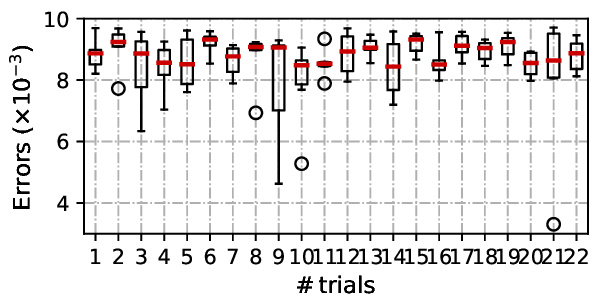}
    \caption{The ultimate reaching errors among different trials.}
    \label{fig:box_plot_err}
\end{figure}

The distribution of the ultimate reaching errors is displayed in the box plot in Fig.~\ref{fig:box_plot_err}. It is noticed that all reaching errors are limited within the error threshold $\varepsilon_N=0.01$\,m. Meanwhile, the overall collision rate $13.64\%$ is comparable with the test collision rates of the IBC-DMP agent shown in Tab.~\ref{tab:collision_rates}. Therefore, the experimental results indicate that the IBC-DMP policy achieves the expected performance as prescribed by the hyperparameter settings, implying the effectiveness of IBC-DMP RL in practical applications.

\section{Discussion}\label{sec:discussion}

This paper uses two important technologies, namely multi-DoF DMP and IBC, to facilitate off-policy RL for robot motion planning. They both can be recognized as the proper encoding of expert knowledge. The multi-DoF DMP is adapted from conventional motion primitives of which the effectiveness has been verified by many experts and peers. From a mathematical point of view, DMP models can be seen as a class of heuristic models dedicated to reducing the dimensionality of a planning problem. The dynamic structure of a DMP model reflects how heuristics are used to restrict the smoothness and inherent stability of the generated trajectories. This represents a different source of expert knowledge other than human demonstrations. IBC, as an adapted version of EBC, provides a soft and elegant way to encode the preference among these two sources of expert knowledge, allowing a flexible knowledge fusion without causing overfitting. As a result, an IBC-DMP RL agent benefits from improved training speed, stability, and generalizability since it always exploits the advantageous aspect of the knowledge. This has been verified by decent results of the experimental studies.

Based on a series of experimental studies, we can answer the questions raised in Sec.~\ref{sec:intro} as follows. 
\begin{itemize}[leftmargin=*]
\item Firstly, using an off-policy algorithm with experience replay (like DDPG) to build a DMP-based robot motion planner is feasible (using IBC-DMP RL). Multi-DoF DMP is suggested for enhanced flexibility. 
\item Secondly, IBC is a decent way of exploiting human demonstrations to promote the learning of a DMP-based RL agent. On the contrary, EBC shows the obvious drawback of overfitting the demonstration data, which may be fatal if the demonstration data does not align with the desired reward. 
\item Thirdly, the usage of multi-DoF DMP and IBC greatly improves the learning efficiency and generalizability of an RL-based robot motion planner if proper hyper-parameters are selected. Unfortunately, similar to other RL-based motion planners, how to determine the best hyper-parameters for an IBC-DMP RL agent is a difficult question and is much dependent on experience.
\end{itemize}

Although the general efficacy of the IBC-DMP framework is validated, it still has limitations. The multi-DoF DMP model has a higher flexibility than the conventional DMP since its virtual force function is dependent on the state of the model. However, this may sacrifice the stability of the DMP model, possibly leading to odd-shaped trajectories. Besides, an IBC-DMP may be over-trained if the number of training episodes is too large, where the training score of the policy decreases or becomes unsteady as the training proceeds longer. This may also be due to the sacrifice of the inherent stability of the multi-DoF DMP. Another limitation of IBC-DMP is that its performance tends to be sensitive to the noise added to actions during the training stage. Further investigations on the correlation between the performance of IBC-DMP and the action noise are needed in future work.

We would also like to highlight that IBC is a promising technology to facilitate training of multi-modal policies or learning from multi-modal datasets. The critical technical point is designing a proper energy function that characterizes multi-modal distributions. Although a deep insight into IBC is not elucidated in this paper since it has mainly focused on the feasibility of using IBC to facilitate RL-based motion planning, we point out that an interesting topic in the future will be fine-tuning or tailoring IBC loss function for different motion planning scenarios, especially those with multi-modal features. The work in~\cite{florence2022implicit} provides a reference for designing experiments for such studies. Moreover, proposing less heuristic and more general loss functions is also an interesting topic to explore in future work.

\section{Conclusion}\label{sec:con}

In this paper, we propose a novel framework to develop RL agents for robot motion-planning tasks based on two promoted methods, namely multi-DoF DMP and IBC. An off-policy RL agent serves as a bridge to flexibly combine the expert knowledge coming from the DMP heuristics and human demonstrations, resulting in an advantageous agent with improved training speed, stability, and generalizability. The efficacy and advantage of an IBC-DMP RL agent over conventional non-IBC agents are verified by experimental studies. In future work, we will focus on improving the stability of the multi-DoF DMP model and the sensitivity of IBC-DMP agents to the action noise.





 
%

\bibliographystyle{IEEEtran}
\bibliography{IEEEabrv, reference.bib}

\begin{thebibliography}{10}
\providecommand{\url}[1]{#1}
\csname url@samestyle\endcsname
\providecommand{\newblock}{\relax}
\providecommand{\bibinfo}[2]{#2}
\providecommand{\BIBentrySTDinterwordspacing}{\spaceskip=0pt\relax}
\providecommand{\BIBentryALTinterwordstretchfactor}{4}
\providecommand{\BIBentryALTinterwordspacing}{\spaceskip=\fontdimen2\font plus
\BIBentryALTinterwordstretchfactor\fontdimen3\font minus \fontdimen4\font\relax}
\providecommand{\BIBforeignlanguage}[2]{{%
\expandafter\ifx\csname l@#1\endcsname\relax
\typeout{** WARNING: IEEEtran.bst: No hyphenation pattern has been}%
\typeout{** loaded for the language `#1'. Using the pattern for}%
\typeout{** the default language instead.}%
\else
\language=\csname l@#1\endcsname
\fi
#2}}
\providecommand{\BIBdecl}{\relax}
\BIBdecl

\bibitem{enayati2022methodical}
A.~M.~S. Enayati, Z.~Zhang, and H.~Najjaran, ``A methodical interpretation of adaptive robotics: Study and reformulation,'' \emph{Neurocomputing}, vol. 512, pp. 381--397, 2022.

\bibitem{wang2021survey}
J.~Wang, T.~Zhang, N.~Ma, Z.~Li, H.~Ma, F.~Meng, and M.~Q.-H. Meng, ``A survey of learning-based robot motion planning,'' \emph{IET Cyber-Systems and Robotics}, vol.~3, no.~4, pp. 302--314, 2021.

\bibitem{kim2015trajectory}
J.-J. Kim and J.-J. Lee, ``Trajectory optimization with particle swarm optimization for manipulator motion planning,'' \emph{IEEE transactions on industrial informatics}, vol.~11, no.~3, pp. 620--631, 2015.

\bibitem{schulman2014motion}
J.~Schulman, Y.~Duan, J.~Ho, A.~Lee, I.~Awwal, H.~Bradlow, J.~Pan, S.~Patil, K.~Goldberg, and P.~Abbeel, ``Motion planning with sequential convex optimization and convex collision checking,'' \emph{The International Journal of Robotics Research}, vol.~33, no.~9, pp. 1251--1270, 2014.

\bibitem{perez2020membrane}
I.~P{\'e}rez-Hurtado, M.~{\'A}. Mart{\'\i}nez-del Amor, G.~Zhang, F.~Neri, and M.~J. P{\'e}rez-Jim{\'e}nez, ``A membrane parallel rapidly-exploring random tree algorithm for robotic motion planning,'' \emph{Integrated Computer-Aided Engineering}, vol.~27, no.~2, pp. 121--138, 2020.

\bibitem{ichter2020learned}
B.~Ichter, E.~Schmerling, T.-W.~E. Lee, and A.~Faust, ``Learned critical probabilistic roadmaps for robotic motion planning,'' in \emph{2020 IEEE International Conference on Robotics and Automation (ICRA)}.\hskip 1em plus 0.5em minus 0.4em\relax IEEE, 2020, pp. 9535--9541.

\bibitem{luis2020online}
C.~E. Luis, M.~Vukosavljev, and A.~P. Schoellig, ``Online trajectory generation with distributed model predictive control for multi-robot motion planning,'' \emph{IEEE Robotics and Automation Letters}, vol.~5, no.~2, pp. 604--611, 2020.

\bibitem{wang2022memory}
Y.~Wang and X.~Guo, ``Memory-based stochastic trajectory optimization for manipulator obstacle avoiding motion planning,'' in \emph{2022 7th Asia-Pacific Conference on Intelligent Robot Systems (ACIRS)}.\hskip 1em plus 0.5em minus 0.4em\relax IEEE, 2022, pp. 188--194.

\bibitem{yang2019survey}
Y.~Yang, J.~Pan, and W.~Wan, ``Survey of optimal motion planning,'' \emph{IET Cyber-Systems and Robotics}, vol.~1, no.~1, pp. 13--19, 2019.

\bibitem{arulkumaran2017deep}
K.~Arulkumaran, M.~P. Deisenroth, M.~Brundage, and A.~A. Bharath, ``Deep reinforcement learning: A brief survey,'' \emph{IEEE Signal Processing Magazine}, vol.~34, no.~6, pp. 26--38, 2017.

\bibitem{bao2022learn}
J.~Bao, G.~Zhang, Y.~Peng, Z.~Shao, and A.~Song, ``Learn multi-step object sorting tasks through deep reinforcement learning,'' \emph{Robotica}, pp. 1--17, 2022.

\bibitem{kulhanek2021visual}
J.~Kulh{\'a}nek, E.~Derner, and R.~Babu{\v{s}}ka, ``Visual navigation in real-world indoor environments using end-to-end deep reinforcement learning,'' \emph{IEEE Robotics and Automation Letters}, vol.~6, no.~3, pp. 4345--4352, 2021.

\bibitem{zhang2022high}
Z.~Zhang, R.~Dershan, A.~M.~S. Enayati, M.~Yaghoubi, D.~Richert, and H.~Najjaran, ``A high-fidelity simulation platform for industrial manufacturing by incorporating robotic dynamics into an industrial simulation tool,'' \emph{IEEE Robotics and Automation Letters}, vol.~7, no.~4, pp. 9123--9128, 2022.

\bibitem{zhao2020sim}
W.~Zhao, J.~P. Queralta, and T.~Westerlund, ``Sim-to-real transfer in deep reinforcement learning for robotics: a survey,'' in \emph{2020 IEEE Symposium Series on Computational Intelligence (SSCI)}.\hskip 1em plus 0.5em minus 0.4em\relax IEEE, 2020, pp. 737--744.

\bibitem{tsurumine2019deep}
Y.~Tsurumine, Y.~Cui, E.~Uchibe, and T.~Matsubara, ``Deep reinforcement learning with smooth policy update: Application to robotic cloth manipulation,'' \emph{Robotics and Autonomous Systems}, vol. 112, pp. 72--83, 2019.

\bibitem{cai2021modular}
M.~Cai, M.~Hasanbeig, S.~Xiao, A.~Abate, and Z.~Kan, ``Modular deep reinforcement learning for continuous motion planning with temporal logic,'' \emph{IEEE Robotics and Automation Letters}, vol.~6, no.~4, pp. 7973--7980, 2021.

\bibitem{yang2021hierarchical}
X.~Yang, Z.~Ji, J.~Wu, Y.-K. Lai, C.~Wei, G.~Liu, and R.~Setchi, ``Hierarchical reinforcement learning with universal policies for multistep robotic manipulation,'' \emph{IEEE Transactions on Neural Networks and Learning Systems}, 2021.

\bibitem{xiong2020comparison}
H.~Xiong, T.~Ma, L.~Zhang, and X.~Diao, ``Comparison of end-to-end and hybrid deep reinforcement learning strategies for controlling cable-driven parallel robots,'' \emph{Neurocomputing}, vol. 377, pp. 73--84, 2020.

\bibitem{voigt2020multi}
F.~Voigt, L.~Johannsmeier, and S.~Haddadin, ``Multi-level structure vs. end-to-end-learning in high-performance tactile robotic manipulation.'' in \emph{CoRL}, 2020, pp. 2306--2316.

\bibitem{da2019survey}
F.~L. Da~Silva and A.~H.~R. Costa, ``A survey on transfer learning for multiagent reinforcement learning systems,'' \emph{Journal of Artificial Intelligence Research}, vol.~64, pp. 645--703, 2019.

\bibitem{stulp2011learning}
F.~Stulp, E.~Theodorou, M.~Kalakrishnan, P.~Pastor, L.~Righetti, and S.~Schaal, ``Learning motion primitive goals for robust manipulation,'' in \emph{2011 IEEE/RSJ International Conference on Intelligent Robots and Systems}.\hskip 1em plus 0.5em minus 0.4em\relax IEEE, 2011, pp. 325--331.

\bibitem{ly2020learning}
A.~O. Ly and M.~Akhloufi, ``Learning to drive by imitation: An overview of deep behavior cloning methods,'' \emph{IEEE Transactions on Intelligent Vehicles}, vol.~6, no.~2, pp. 195--209, 2020.

\bibitem{fang2019survey}
B.~Fang, S.~Jia, D.~Guo, M.~Xu, S.~Wen, and F.~Sun, ``Survey of imitation learning for robotic manipulation,'' \emph{International Journal of Intelligent Robotics and Applications}, vol.~3, no.~4, pp. 362--369, 2019.

\bibitem{ravichandar2020recent}
H.~Ravichandar, A.~S. Polydoros, S.~Chernova, and A.~Billard, ``Recent advances in robot learning from demonstration,'' \emph{Annual review of control, robotics, and autonomous systems}, vol.~3, pp. 297--330, 2020.

\bibitem{rajeswaran2017learning}
A.~Rajeswaran, V.~Kumar, A.~Gupta, G.~Vezzani, J.~Schulman, E.~Todorov, and S.~Levine, ``Learning complex dexterous manipulation with deep reinforcement learning and demonstrations,'' \emph{arXiv preprint arXiv:1709.10087}, 2017.

\bibitem{tian2021learning}
Y.~Tian, X.~Cao, K.~Huang, C.~Fei, Z.~Zheng, and X.~Ji, ``Learning to drive like human beings: A method based on deep reinforcement learning,'' \emph{IEEE Transactions on Intelligent Transportation Systems}, 2021.

\bibitem{nair2018overcoming}
A.~Nair, B.~McGrew, M.~Andrychowicz, W.~Zaremba, and P.~Abbeel, ``Overcoming exploration in reinforcement learning with demonstrations,'' in \emph{2018 IEEE international conference on robotics and automation (ICRA)}.\hskip 1em plus 0.5em minus 0.4em\relax IEEE, 2018, pp. 6292--6299.

\bibitem{gupta2021reinforcement}
K.~Gupta, ``Reinforcement learning in complex environments with locally trained na{\"\i}ve agents,'' Ph.D. dissertation, University of British Columbia, 2021.

\bibitem{florence2022implicit}
P.~Florence, C.~Lynch, A.~Zeng, O.~A. Ramirez, A.~Wahid, L.~Downs, A.~Wong, J.~Lee, I.~Mordatch, and J.~Tompson, ``Implicit behavioral cloning,'' in \emph{Conference on Robot Learning}.\hskip 1em plus 0.5em minus 0.4em\relax PMLR, 2022, pp. 158--168.

\bibitem{stulp2012reinforcement}
F.~Stulp, E.~A. Theodorou, and S.~Schaal, ``Reinforcement learning with sequences of motion primitives for robust manipulation,'' \emph{IEEE Transactions on robotics}, vol.~28, no.~6, pp. 1360--1370, 2012.

\bibitem{schaal2006dynamic}
S.~Schaal, ``Dynamic movement primitives-a framework for motor control in humans and humanoid robotics,'' in \emph{Adaptive motion of animals and machines}.\hskip 1em plus 0.5em minus 0.4em\relax Springer, 2006, pp. 261--280.

\bibitem{stulp2011hierarchical}
F.~Stulp and S.~Schaal, ``Hierarchical reinforcement learning with movement primitives,'' in \emph{2011 11th IEEE-RAS International Conference on Humanoid Robots}.\hskip 1em plus 0.5em minus 0.4em\relax IEEE, 2011, pp. 231--238.

\bibitem{cohen2021motion}
Y.~Cohen, O.~Bar-Shira, and S.~Berman, ``Motion adaptation based on learning the manifold of task and dynamic movement primitive parameters,'' \emph{Robotica}, vol.~39, no.~7, pp. 1299--1315, 2021.

\bibitem{kulvicius2011joining}
T.~Kulvicius, K.~Ning, M.~Tamosiunaite, and F.~Worg{\"o}tter, ``Joining movement sequences: Modified dynamic movement primitives for robotics applications exemplified on handwriting,'' \emph{IEEE Transactions on Robotics}, vol.~28, no.~1, pp. 145--157, 2011.

\bibitem{hogan2012dynamic}
N.~Hogan and D.~Sternad, ``Dynamic primitives of motor behavior,'' \emph{Biological cybernetics}, vol. 106, no.~11, pp. 727--739, 2012.

\bibitem{li2021reinforcement}
A.~Li, Z.~Liu, W.~Wang, M.~Zhu, Y.~Li, Q.~Huo, and M.~Dai, ``Reinforcement learning with dynamic movement primitives for obstacle avoidance,'' \emph{Applied Sciences}, vol.~11, no.~23, p. 11184, 2021.

\bibitem{liang2021dynamic}
Y.~Liang, W.~Li, Y.~Wang, R.~Xiong, Y.~Mao, and J.~Zhang, ``Dynamic movement primitive based motion retargeting for dual-arm sign language motions,'' in \emph{2021 IEEE International Conference on Robotics and Automation (ICRA)}.\hskip 1em plus 0.5em minus 0.4em\relax IEEE, 2021, pp. 8195--8201.

\bibitem{yuan2022hierarchical}
Y.~Yuan, Z.~L. Yu, L.~Hua, Y.~Cheng, J.~Li, and X.~Sang, ``Hierarchical dynamic movement primitive for the smooth movement of robots based on deep reinforcement learning,'' \emph{Applied Intelligence}, pp. 1--18, 2022.

\bibitem{saveriano2023dynamic}
M.~Saveriano, F.~J. Abu-Dakka, A.~Kramberger, and L.~Peternel, ``Dynamic movement primitives in robotics: A tutorial survey,'' \emph{The International Journal of Robotics Research}, vol.~42, no.~13, pp. 1133--1184, 2023.

\bibitem{sun2021motion}
H.~Sun, W.~Zhang, R.~Yu, and Y.~Zhang, ``Motion planning for mobile robots—focusing on deep reinforcement learning: A systematic review,'' \emph{IEEE Access}, vol.~9, pp. 69\,061--69\,081, 2021.

\bibitem{zhou2022review}
C.~Zhou, B.~Huang, and P.~Fr{\"a}nti, ``A review of motion planning algorithms for intelligent robots,'' \emph{Journal of Intelligent Manufacturing}, vol.~33, no.~2, pp. 387--424, 2022.

\bibitem{yu2022intelligent}
L.~Yu, J.~Luo, and K.~Zhou, ``An intelligent robot motion planning method and application via lppo in unknown environment,'' in \emph{2022 12th International Conference on CYBER Technology in Automation, Control, and Intelligent Systems (CYBER)}.\hskip 1em plus 0.5em minus 0.4em\relax IEEE, 2022, pp. 265--270.

\bibitem{ying2022trajectory}
F.~Ying, H.~Liu, R.~Jiang, and X.~Yin, ``Trajectory generation for multiprocess robotic tasks based on nested dual-memory deep deterministic policy gradient,'' \emph{IEEE/ASME Transactions on Mechatronics}, 2022.

\bibitem{fan2018surreal}
L.~Fan, Y.~Zhu, J.~Zhu, Z.~Liu, O.~Zeng, A.~Gupta, J.~Creus-Costa, S.~Savarese, and L.~Fei-Fei, ``Surreal: Open-source reinforcement learning framework and robot manipulation benchmark,'' in \emph{Conference on Robot Learning}.\hskip 1em plus 0.5em minus 0.4em\relax PMLR, 2018, pp. 767--782.

\bibitem{naughton2021elastica}
N.~Naughton, J.~Sun, A.~Tekinalp, T.~Parthasarathy, G.~Chowdhary, and M.~Gazzola, ``Elastica: A compliant mechanics environment for soft robotic control,'' \emph{IEEE Robotics and Automation Letters}, vol.~6, no.~2, pp. 3389--3396, 2021.

\bibitem{hou2017novel}
Y.~Hou, L.~Liu, Q.~Wei, X.~Xu, and C.~Chen, ``A novel ddpg method with prioritized experience replay,'' in \emph{2017 IEEE international conference on systems, man, and cybernetics (SMC)}.\hskip 1em plus 0.5em minus 0.4em\relax IEEE, 2017, pp. 316--321.

\bibitem{luo2020dynamic}
J.~Luo and H.~Li, ``Dynamic experience replay,'' in \emph{Conference on robot learning}.\hskip 1em plus 0.5em minus 0.4em\relax PMLR, 2020, pp. 1191--1200.

\bibitem{choi2020robotic}
J.~Choi, H.~Kim, Y.~Son, C.-W. Park, and J.~H. Park, ``Robotic behavioral cloning through task building,'' in \emph{2020 International Conference on Information and Communication Technology Convergence (ICTC)}.\hskip 1em plus 0.5em minus 0.4em\relax IEEE, 2020, pp. 1279--1281.

\bibitem{tamizi2024end}
M.~G. Tamizi, H.~Honari, A.~Nozdryn-Plotnicki, and H.~Najjaran, ``End-to-end deep learning-based framework for path planning and collision checking: bin-picking application,'' \emph{Robotica}, vol.~42, no.~4, pp. 1094--1112, 2024.

\bibitem{farag2018behavior}
W.~Farag and Z.~Saleh, ``Behavior cloning for autonomous driving using convolutional neural networks,'' in \emph{2018 International Conference on Innovation and Intelligence for Informatics, Computing, and Technologies (3ICT)}.\hskip 1em plus 0.5em minus 0.4em\relax IEEE, 2018, pp. 1--7.

\bibitem{chen2017robot}
C.~Chen, C.~Yang, C.~Zeng, N.~Wang, and Z.~Li, ``Robot learning from multiple demonstrations with dynamic movement primitive,'' in \emph{2017 2nd International Conference on Advanced Robotics and Mechatronics (ICARM)}.\hskip 1em plus 0.5em minus 0.4em\relax IEEE, 2017, pp. 523--528.

\bibitem{li2023prodmp}
G.~Li, Z.~Jin, M.~Volpp, F.~Otto, R.~Lioutikov, and G.~Neumann, ``Prodmp: A unified perspective on dynamic and probabilistic movement primitives,'' \emph{IEEE Robotics and Automation Letters}, vol.~8, no.~4, pp. 2325--2332, 2023.

\bibitem{gupta2022exploiting}
K.~Gupta and H.~Najjaran, ``Exploiting abstract symmetries in reinforcement learning for complex environments,'' in \emph{2022 International Conference on Robotics and Automation (ICRA)}.\hskip 1em plus 0.5em minus 0.4em\relax IEEE, 2022, pp. 3631--3637.

\bibitem{kumar2021should}
A.~Kumar, J.~Hong, A.~Singh, and S.~Levine, ``{Should} {I} run offline reinforcement learning or behavioral cloning?'' in \emph{International Conference on Learning Representations}, 2021.

\bibitem{wang2023diffusion}
Z.~Wang, J.~J. Hunt, and M.~Zhou, ``Diffusion policies as an expressive policy class for offline reinforcement learning,'' \emph{arXiv/2208.06193}, 2023.

\bibitem{shafiullah2022behavior}
N.~M. Shafiullah, Z.~Cui, A.~A. Altanzaya, and L.~Pinto, ``Behavior transformers: Cloning $ k $ modes with one stone,'' \emph{Advances in neural information processing systems}, vol.~35, pp. 22\,955--22\,968, 2022.

\bibitem{paranjape2015motion}
A.~A. Paranjape, K.~C. Meier, X.~Shi, S.-J. Chung, and S.~Hutchinson, ``Motion primitives and 3d path planning for fast flight through a forest,'' \emph{The International Journal of Robotics Research}, vol.~34, no.~3, pp. 357--377, 2015.

\bibitem{SpinningUp2018}
J.~Achiam, ``{Spinning Up in Deep Reinforcement Learning},'' 2018.

\bibitem{Jocher_YOLO_by_Ultralytics_2023}
\BIBentryALTinterwordspacing
G.~Jocher, A.~Chaurasia, and J.~Qiu, ``{YOLO by Ultralytics},'' Jan. 2023. [Online]. Available: \url{https://github.com/ultralytics/ultralytics}
\BIBentrySTDinterwordspacing

\bibitem{opencv_library}
G.~Bradski, ``{The OpenCV Library},'' \emph{Dr. Dobb's Journal of Software Tools}, 2000.

\end{thebibliography}


 

\balance

\end{document}